\def\inclapp{1}
\def\viewchanges{1}
\def\viewauthors{1}
\def\addackn{1}
\def\twocol{1}
\def\broaderimpact{0}
\def\preprint{1}
\def\aistats{0}

\if\aistats1	
	\documentclass[twoside]{article}
\else
	\documentclass{article}
\fi

\usepackage{microtype}
\usepackage{booktabs} 
\usepackage{amsmath}
\usepackage{amssymb}
\usepackage[normalem]{ulem}
\usepackage{caption}
\usepackage[normalem]{ulem}
\usepackage{amsthm}
\usepackage{bbm}

\usepackage[round]{natbib}

\usepackage{graphicx,wrapfig,lipsum}
\usepackage{subfigure}
\usepackage{algorithmic}
\usepackage{algorithm}
\usepackage{wrapfig}

\usepackage{xr}
\newtheorem{theorem}{Theorem}[section]

\newtheorem{lem}[theorem]{Lemma}

\newtheorem{rem}[theorem]{Remark}
\newtheorem{cor}[theorem]{Corollary}

\theoremstyle{definition}
\newtheorem{examplex}[theorem]{Example}
\newenvironment{example}
  {\pushQED{\qed}\examplex}
  {\popQED\endexamplex}

\let\P\undefined%
\newcommand{\P}{\mathbb{P}}%
\let\E\undefined%
\newcommand{\E}{\mathbb{E}}%
\let\R\undefined%
\newcommand{\R}{\mathbb{R}}%
\let\N\undefined%
\newcommand{\N}{\mathbb{N}}%

\usepackage[dvipsnames]{xcolor}

\let\del\undefined
\let\com\undefined

\if\viewchanges1
	
	\newcommand{\del}{\textcolor{red}}
	\newcommand{\com}{\textcolor{orange}}
	
\else
	
	\newcommand{\del}[1]{}
	\newcommand{\com}[1]{}
	
\fi

\def\Title{Local Lipschitz Bounds of Deep Neural Networks}

\if\twocol1
	\usepackage[colorlinks, citecolor=blue]{hyperref}
	
	\if\aistats1
		\usepackage{aistats2021}
	\else
	     \usepackage[accepted]{icml2020}
	     \icmltitlerunning{\Title}
	\fi
	
\else
	\usepackage[colorlinks, citecolor=blue, linkcolor=magenta]{hyperref}       
	\usepackage{algorithm}
	\usepackage{algorithmic}

	\if\preprint0
		\usepackage{neurips_2020}
	\else
	     \usepackage[preprint]{neurips_2020}
	\fi
	
	\usepackage{natbib}
\fi

\begin{document}

\if\twocol1
	\if\aistats1
	   \twocolumn[\aistatstitle{\Title}
	   
		\aistatsauthor{ Author 1 \And Author 2 \And  Author 3 }
	
		\aistatsaddress{ Institution 1 \And  Institution 2 \And Institution 3 } ]
	\else

		\twocolumn[
			\icmltitle{\Title}
		
		
		
		
		\begin{icmlauthorlist}
		\icmlauthor{Calypso Herrera}{ETH}
		\icmlauthor{Florian Krach}{ETH}
		\icmlauthor{Josef Teichmann}{ETH}
		\end{icmlauthorlist}
		\icmlaffiliation{ETH}{Department of Mathematics, ETH Z\"urich, Switzerland}
		\icmlcorrespondingauthor{}{firstname.lastname@math.ethz.ch}
		
		\icmlkeywords{Machine Learning, ICML}
		
		\vskip 0.3in
		]
	
		
		\printAffiliationsAndNotice{}  
	\fi
\else
	\title{\Title}
	\if\viewauthors1
		\author{%
		  Calypso Herrera\\
		  Department of Mathematics\\
		  ETH Zurich, Switzerland \\
		  \texttt{calypso.herrera@math.ethz.ch} \\
		   \And
		  Florian Krach\\
		  Department of Mathematics\\
		  ETH Zurich, Switzerland \\
		  \texttt{florian.krach@math.ethz.ch} \\
		   \And
		  Josef Teichmann\\
		  Department of Mathematics\\
		  ETH Zurich, Switzerland \\
		  \texttt{josef.teichmann@math.ethz.ch} 
		}
		\maketitle
	\else
		\maketitle
	\fi
\fi

\begin{abstract}
The Lipschitz constant is an important quantity that arises in analysing the convergence of gradient-based optimization methods. It is generally unclear how to estimate the Lipschitz constant of a complex model. Thus, this paper studies an important problem that may be useful to the broader area of non-convex optimization.
The main result provides a local upper bound on the Lipschitz constants of a multi-layer feed-forward neural network and its gradient. Moreover, lower bounds are established as well, which are used to show that it is impossible to derive global upper bounds for the Lipschitz constants.
In contrast to previous works, we compute the Lipschitz constants with respect to the network parameters and not with respect to the inputs. These constants are needed for the theoretical description of many step size schedulers of gradient based optimization schemes and their convergence analysis. 
The idea is both simple and effective. 
The results are extended to a generalization of neural networks, continuously deep neural networks, which are described by controlled ODEs.

\end{abstract}

\section{Introduction}
 
The training of a neural network can be summarized as an optimization problem which consists of
making steps towards extrema of a loss function.
Variants of the stochastic gradient descent (SGD) are generally used to solve this problem. They give surprisingly good results, even though the objective function is not convex in most cases.
The adaptive gradient methods are a state-of-the-art variation of SGD. In particular, AdaGrad \citep{Duchi:2011:ASM:1953048.2021068}, RMSProp \citep{Tieleman2012}, and ADAM \citep{kingma2014adam} are widely used methods to train neural networks \citep{Melis2017OnTS, pmlr-v37-xuc15}.
In most of the SGD methods, the rate of convergence depends on the Lipschitz constant of the gradient of the loss function with respect the parameters \citep{Reddi2018,Li_Orabona_2019}. Therefore, it is essential to have an  upper bound estimate on the Lipschitz constant in order to get a better understanding of the convergence and to be able to set an appropriate step-size. 

In this paper, we provide a general and efficient estimate for upper bounds on the Lipschitz constant of the gradient of any loss function applied to a feed-forward fully connected DNN with respect to the parameters. Naturally, this estimate depends on the architecture of the DNN (i.e. the activation function, the depth of the NN, the size of the layers) as well as on the norm of the input and on the loss function. 
In two examples we also establish lower bounds on the Lipschitz constants, confirming that worst-case estimates grow exponentially in the number of layers.

As a concrete application, we show how our estimate can be used to set the (hyper-parameters of the) step size of the AdaGrad \citep{Li_Orabona_2019} SGD method, such that convergence of this optimization scheme is guaranteed (in expectation). In particular, the convergence rate of AdaGrad with respect to the Lipschitz estimate of the gradient of the loss function can be calculated.

In addition, we provide Lipschitz estimates for any neural network that can be represented as solution of a controlled ordinary differential equation (controlled ODE) \citep{cuchiero2019deep}. 
This includes classical DNN as well as \emph{continuously deep neural networks}, like \emph{neural ODE} \citep{Chen2018}, \emph{ODE-RNN} \citep{DBLP:journals/corr/abs-1907-03907}, \emph{GRU-ODE-Bayes} \citep{Brouwer2019GRUODEBayesCM} and \emph{neural SDE} \citep{DBLP:journals/corr/abs-1906-02355, DBLP:journals/corr/abs-1905-09883, NJSDE}. Therefore, having such a general Lipschitz estimate allows us to cover a wide range of architectures and to study their convergence behaviour. Moreover, controlled ODE can provide us with neural-network based parametrized families of invertible functions (cf. \cite{cuchiero2019deep}), including in particular feed-forward neural networks. 

\section{Related work}
Very recently, several estimates of the Lipschitz constants of neural networks were proposed \citep{bartlett2017spectrally, Scaman2018LipschitzRO, Combettes2019LipschitzCF, Fazlyab2019EfficientAA, Jin2018StabilitycertifiedRL, Raghunathan2018CertifiedDA, pmlr-v80-arora18b, Latorre2020LipschitzCE}. In contrast to our work, those estimates are upper bounds on the Lipschitz constants of neural networks \emph{with respect to the inputs} and not with respect to the parameters as we provide here. 
Those works are mainly concerned with the sensitivity of neural networks to their inputs, while our main goal is to provide bounds on the Lipschitz constants of general DNNs and their gradients, which are needed to study their convergence properties.
\citep{zou2018stochastic, li2018tighter, cao2019generalization, allen2019convergence} give bounds on the Lipschitz constant of a DNN but not on the Lipschitz constant of the gradient of the DNN as we do.

In the classical setting, results similar to our work were given in \cite{baes2019lowrank} for one specific loss function. In comparison, we provide in the classical setting a simplified proof.
To the best of our knowledge, neither for the classical setting of deep feed-forward fully connected neural networks, nor for the controlled ODE framework, general estimates of the Lipschitz constants \emph{with respect to the parameters} are available.

Continuously deep neural networks are already used in practice. 
The \emph{neural ODE} introduced in \cite{Chen2018}, is an example of such a continuously deep neural network that can be described in our framework by \eqref{eq:solved controlled ODE}, when choosing $d=1$ and $u_1(t)=t$, i.e. $du_1(t) = dt$. However, our framework allows to describe more general architectures, which combine jumps (as occurring in Example \ref{exa:CODE-NN equivalence}) with continuous evolutions as in \cite{Chen2018}. One example of such an architecture is the ODE-RNN introduced in \cite{DBLP:journals/corr/abs-1907-03907}.
Furthermore, allowing $u_i$ to be semimartingales instead of deterministic processes of finite variation, \emph{neural SDE} models as described e.g. in \cite{DBLP:journals/corr/abs-1906-02355, NJSDE, Peluchetti2019InfinitelyDN, DBLP:journals/corr/abs-1905-09883} are covered by our framework \eqref{eq:solved controlled ODE}.

\section{Ordinary deep neural network setting}\label{sec:Ordinary deep neural network setting}
\subsection{Problem setup}\label{sec:Problem set-up}
The norm we shall use in the sequel is a natural extension of the standard
Frobenius norm to finite lists of matrices of diverse sizes. Specifically,
for any $\gamma\in\mathbb{N}$, $m_{1},\ldots,m_{\gamma},n_{1},\ldots,n_{\gamma}\in\mathbb{N}$,
and $(M^{1},\ldots,M^{\gamma})\in\mathbb{R}^{m_{1}\times n_{1}}\times\dots\times\mathbb{R}^{m_{\gamma}\times n_{\gamma}}$,
we let 
\begin{multline}\label{eq:Euclidean-like norm-Intro}
\left\Vert \left(M^{1},\dots,M^{\gamma}\right)\right\Vert^2 :=
\sum_{k=1}^\gamma  \sum_{i=1}^{m_{k}}\sum_{j=1}^{n_{k}}\left(M_{i,j}^{k}\right)^{2}  .
\end{multline}
Furthermore we use the maximum norm, defined as
\begin{equation*}
\lVert \left(M^{1},\dots,M^{\gamma}\right) \rVert_{\infty} := \max_{i, j, k} \{ \lvert M_{i,j}^k \rvert \}.
\end{equation*}

Consider positive integers $\ell_{u}$ for $ u = 0,\dots,m+1$. We construct a \emph{deep neural network} (DNN) with $m$ layers of $\ell_u$, $u\in \{1,\dots,m\}$ neurons, each with an (activation) function $\tilde\sigma_u:\mathbb{R}\to \R$, such that there exist $ \sigma_{\max}, \sigma'_{\max}, \sigma''_{\max}  > 0$, so that for all $ u \in \{1, \dots, m\}$ and all $x \in \mathbb{R}$ we have $\vert \tilde\sigma_u(x) \vert  \leq  \sigma_{\max}$, $\vert \tilde\sigma_u'(x) \vert  \leq  \sigma'_{\max}$ and $\vert \tilde\sigma_u''(x) \vert  \leq  \sigma''_{\max}$. 
This assumption is met by the classical sigmoid and $\tanh$ functions, but excludes the popular ReLU activation function. However, our main results of this section can easily be extended to allow for ReLU as well, as outlined in Section~\ref{sec:Activation functions}.
For each $u \in \{1,\dots,m+1\}$, let $A^{(u)}=\left[A^{(u)}_{i,j}\right]_{i,j}\in\mathbb{R}^{\ell_u\times \ell_{u-1}}$ be the
\emph{weights} and $b^{(u)}=\left[b^{(u)}_{i}\right]_{i}\in\mathbb{R}^{\ell_u}$ be the
\emph{bias}. Let $\theta_u = \left(A^{(u)},b^{(u)}  \right)$ and define for every $u \in \{1,\dots,m+1\}$
\if\twocol1
	\begin{align*}
	&f_{\theta_u}:\mathbb{R}^{{\ell_{u-1}}}&\to\mathbb{R}^{\ell_u}, \quad x&\mapsto A^{(u)} x+b^{(u)},\\
	&\sigma_{u}:\mathbb{R}^{\ell_u}&\to\mathbb{R}^{\ell_u}, \quad x&\mapsto\left(\tilde{\sigma}_u(x_{1}),\dots,\tilde{\sigma}_u(x_{\ell_u})\right)^{\top} .
	\end{align*}
\else
	\begin{align*}
	&f_{\theta_u}:\mathbb{R}^{{\ell_{u-1}}}\to\mathbb{R}^{\ell_u}, \quad x\mapsto A^{(u)} x+b^{(u)},\\
	&\sigma_{u}:\mathbb{R}^{\ell_u}\to\mathbb{R}^{\ell_u}, \quad x\mapsto\left(\tilde{\sigma}_u(x_{1}),\dots,\tilde{\sigma}_u(x_{\ell_u})\right)^{\top} .
	\end{align*}
\fi

We denote for every $u \in \{1,\dots,m+1\}$ the parameters  $\Theta_{u}:=(\theta_1, \dots, \theta_{u}) $, and by a slight abuse of notation, considering $\theta_u$ and $\Theta_u$ as flattened vectors, we write $\theta_u \in \R^{\tilde{d}_u}$ and $\Theta_u \in \R^{d_u}$. Moreover, we define $\Omega \subset \R^{d_{m+1}}$ as the set of possible neural  \emph{network parameters}. 
Then we define the $m$-layered feed-forward neural network as the function
\if\twocol1
	\begin{eqnarray} \label{eq:NN definition}
	\mathcal{N}_{\Theta_{m+1}}:\mathbb{R}^{\ell_0} 
	&\to & \mathbb{R}^{\ell_{m+1}}\\ 	z  &\mapsto &  f_{\theta_{m+1}} \circ \sigma_m \circ f_{\theta_{m}} \circ \dots \circ \sigma_1\circ f_{\theta_1}(z)\,. \nonumber
	\end{eqnarray}
\else
	\begin{equation}\label{eq:NN definition}
	\mathcal{N}_{\Theta_{m+1}}:\mathbb{R}^{\ell_0} 
	\to \mathbb{R}^{\ell_{m+1}}, \quad	
	z  \mapsto   f_{\theta_{m+1}} \circ \sigma_m \circ f_{\theta_{m}} \circ \dots \circ \sigma_1\circ f_{\theta_1}(z)\,.
	\end{equation}
\fi
By $\mathbb{L}_{m+1}:= d_{m+1} = \sum_{u=1}^{m+1} \left(\ell_u \ell_{u-1} + \ell_u\right)$ 
we denote the number of trainable parameters of $\mathcal{N}_{\Theta_{m+1}}$. 

We now assume that there exists a (possibly infinite) set of possible training samples $\mathcal{Z} \subset \R^{\ell_0} \times \R^{k}$, for $k \in \N$, equipped with a sigma algebra $\mathcal{A}(\mathcal{Z})$ and a probability measure $\P$, the distribution of the training samples. Let $Z \sim \P$ be a random variable following this distribution. We use the notation $Z = (Z_x, Z_y) = (\operatorname{proj}_x(Z), \operatorname{proj}_y(Z))$ to emphasize the two components of a training sample $Z$. In a standard supervised learning setup we have $k = \ell_{m+1}$, where $Z_x \in \R^{\ell_0}$ is the \emph{input} and $Z_y \in \R^{k}$ is the \emph{target}.
However, we also allow any other setup including $k = 0$, corresponding to training samples consisting only of the input, i.e. an unsupervised setting.
Let
\begin{equation*}
g:\mathbb{R}^{{\ell_{m+1}}} \times \R^{k} \to \mathbb{R},\quad (x,y) \mapsto g(x,y),
\end{equation*}
be a function which is twice differentiable in the first component. We assume there exist $g_{\max}^{\prime}, g_{\max}^{\prime \prime} > 0$ such that for all $(x,y) \in \mathbb{R}^{{\ell_{m+1}}} \times \R^{k}$ we have $ \lVert \tfrac{\partial}{\partial x} g(x, y) \rVert \leq g_{\max}^{\prime}$ and $ \lVert \tfrac{\partial^2}{\partial x^2} g(x, y) \rVert \leq g_{\max}^{\prime \prime}$. We use $g$ to define the cost function, given one training sample $\zeta := (\zeta_x,\zeta_y) \in\mathbb{R}^{\ell_0} \times \R^{k}$, as
\if\twocol1
	\begin{align*}
	\varphi: \mathbb{R}^{\mathbb{L}_{m+1}} \times (\mathbb{R}^{\ell_0} \times \R^{k}) & \to \mathbb{R} \\
		(\Theta_{m+1}, \zeta) & \mapsto g\left(\mathcal{N}_{\Theta_{m+1}}(\zeta_x), \zeta_y \right).
	\end{align*}
\else
	\begin{equation*}
	\varphi: \mathbb{R}^{\mathbb{L}_{m+1}} \times (\mathbb{R}^{\ell_0} \times \R^{k})  \to \mathbb{R}, \quad
		(\Theta_{m+1}, \zeta)  \mapsto g\left(\mathcal{N}_{\Theta_{m+1}}(\zeta_x), \zeta_y \right).
	\end{equation*}
\fi
Then we define the \emph{cost function} (interchangeably called \emph{objective} or \emph{loss function}) as 
\begin{align*}
\Phi : \R^{\mathbb{L}_{m+1}} \to \R, \; \Theta_{m+1} \mapsto \E[ \varphi(\Theta_{m+1}, Z) ],
\end{align*}
where we denote by $\E$ the expectation with respect to $\P$.
We note that for a finite set of training samples $\mathcal{Z} = \lbrace \zeta_1, \dotsc, \zeta_N \rbrace$, with equal probabilities (Laplace probability model) we obtain the standard neural network objective function $\Phi(\Theta_{m+1}) = \tfrac{1}{N} \sum_{i=1}^N \varphi(\Theta_{m+1}, \zeta_i)$.

\subsection{Main results}\label{sec:Main results}
The following theorems show that under standard assumptions for neural network training, the neural network $\mathcal{N}$, as well as the cost function $\Phi$, are Lipschitz continuous with Lipschitz continuous gradients with respect to the parameters $\Theta_{m+1}$. We explicitly calculate upper bounds on the Lipschitz constants. Furthermore, we apply these results to infer a bound on the convergence rate to \emph{a stationary point} of the cost function. The proofs are given in Appendix \ref{sec:Proofs in the ordinary DNN setting}. 

To simplify the notation, we define the following functions for a given training sample $(\zeta_x, \zeta_y) \in \mathcal{Z}$ and $2 \leq u \leq m$:
\if\twocol1
	\begin{align*}
	&N_1 :  &\R^{d_1}& &\to& \R^{\ell_1},\quad \Theta_1& \mapsto & \,\sigma_1 \circ f_{\theta_1} (\zeta_x), \\
	&N_u :  &\R^{d_u}& &\to & \R^{\ell_u},\quad \Theta_u& \mapsto& \, \sigma_u \circ f_{\theta_u} ( N_{u-1} (\Theta_{u-1}) ), \\
	&N : &\Omega& &\to& \R^{\ell_{m+1}}, \quad \Theta& \mapsto& \, \mathcal{N}_{\Theta}(\zeta_x).
	\end{align*}
\else
	\begin{equation*}
	\begin{split}
	N_1 :  &\R^{d_1} \to \R^{\ell_1},\quad \Theta_1 \mapsto \,\sigma_1 \circ f_{\theta_1} (\zeta_x), \\
	N_u :  &\R^{d_u} \to  \R^{\ell_u},\quad \Theta_u \mapsto \, \sigma_u \circ f_{\theta_u} ( N_{u-1} (\Theta_{u-1}) ), \\
	N : &\Omega \to \R^{\ell_{m+1}}, \quad \Theta \mapsto \, \mathcal{N}_{\Theta}(\zeta_x).
	\end{split}
	\end{equation*}
\fi

First we derive upper bounds on the Lipschitz constants of the neural network $N$ and its gradient.
\begin{theorem}\label{rem:lipschitz of NN} \label{thm:lipschitz of NN}
We assume that the space of network parameters $\Omega$ is non-empty, open and bounded in the maximum norm, that is, there exists some $0 < B_{\Omega} < \infty$ such that for every $\Theta \in \Omega$ we have $\lVert \Theta \rVert_{\infty} < B_{\Omega}$. For any fixed training sample $(\zeta_x, \zeta_y) \in \mathcal{Z}$ we set $S := \lVert \zeta_x \rVert$.
Then, for $1\leq u \leq m$, each $N_u$ and its gradient $\nabla N_u$ are Lipschitz continuous with constants $L_{N_u}$ and $L_{\nabla N_u}$ and uniformly bounded with constants $B_{N_u}$, $B_{\nabla N_{u}} = L_{N_u}$, which can be upper bounded as follows:
\if\twocol1
	\begin{align*}
	B_{N_1} &= \sqrt{\ell_1} \sigma_{\max}, \\
	L_{N_1} &= \sigma'_{\max} \sqrt{S^2 + 1}, \\
	L_{\nabla N_1} &=  \sigma''_{\max} \sqrt{(S^2 + 1) (3 S^2 + 2) },
	\end{align*}
	\begin{align*}
	D_u &= \sqrt{ \ell_{u-1} \ell_u } B_{\Omega},\\
	B_{N_u} &= \sqrt{\ell_u} \sigma_{\max}, \\
	L_{N_u} &= \sigma'_{\max} \sqrt{D_u^2 L_{N_{u-1}}^2 + B_{N_{u-1}}^2 + 1}, \\
	L_{\nabla N_u} &= \sqrt{\alpha_{u} + \beta_u}, \\
	\alpha_u &= \max\lbrace 3 L_{N_{u-1}}^2  ((\sigma'_{\max})^2 \ell_u + (\sigma''_{\max})^2 D_u^2 B_{N_{u-1}}^2) \\
			& \quad \quad \quad \quad + 2 (\sigma''_{\max})^2 D_u^2 L_{N_{u-1}}^2 \, , \\ 
			& \quad \quad \quad \; \; (\sigma''_{\max})^2 ( B_{N_{u-1}}^2 + 1 )(3 B_{N_{u-1}}^2 + 2 ) \rbrace,  \\
	\beta_u &= (\ell_u \sigma'_{\max} D_u L_{\nabla N_{u-1}} +  \sigma''_{\max} D_u^2  L_{N_{u-1}}^2)^2 \\
			& \quad + L_{N_{u-1}}^2 \big( \ell_u \sigma'_{\max} + D_u \sigma''_{\max} (B_{N_{u-1}}^2 + 1)^{1/2} \big)^2.
	\end{align*}
\else
	\begin{align*}
	L_{N_1} &= \sigma'_{\max} \sqrt{S^2 + 1}, \quad
	B_{N_1} = \sqrt{\ell_1} \sigma_{\max}, \quad
	L_{\nabla N_1} =  \sigma''_{\max} \sqrt{(S^2 + 1) (3 S^2 + 2) }, \\
	\end{align*}
	\begin{align*}
	D_u &= \sqrt{ \ell_{u-1} \ell_u } B_{\Omega},\\
	L_{N_u} &= \sigma'_{\max} \sqrt{D_u^2 L_{N_{u-1}}^2 + B_{N_{u-1}}^2 + 1}, \quad
	B_{N_u} = \sqrt{\ell_u} \sigma_{\max}, \quad
	L_{\nabla N_u} = \sqrt{\alpha_{u} + \beta_u}, \\
	\alpha_u &= \max\lbrace 3 L_{N_{u-1}}^2  ((\sigma'_{\max})^2 \ell_u + (\sigma''_{\max})^2 D_u^2 B_{N_{u-1}}^2) + 2 (\sigma''_{\max})^2 D_u^2 L_{N_{u-1}}^2 \, , \\ 
			& \quad \quad \quad \; \; (\sigma''_{\max})^2 ( B_{N_{u-1}}^2 + 1 )(3 B_{N_{u-1}}^2 + 2 ) \rbrace,  \\
	\beta_u &= (\ell_u \sigma'_{\max} D_u L_{\nabla N_{u-1}} + \sigma''_{\max} D_u^2  L_{N_{u-1}}^2)^2 + L_{N_{u-1}}^2 \big( \ell_u \sigma'_{\max} + D_u \sigma''_{\max} (B_{N_{u-1}}^2 + 1)^{1/2} \big)^2 .
	\end{align*}
\fi
Furthermore, the function $N$ and its gradient $\nabla N$ are Lipschitz continuous with constant $L_N$ and $L_{\nabla N}$. This also implies that $\nabla N$ is uniformly bounded by $B_{\nabla N} = L_N$ and these constants can be estimated by
\if\twocol1
	\begin{align*}
	L_N &= B_{\nabla N } = \sqrt{D_u^2 L_{N_{m}}^2 + B_{N_{m}}^2 + 1}, \\
	L_{\nabla N} &= \sqrt{3 L_{N_{m}}^2 \ell_{m+1} + \ell_{m+1}^2  D_u^2 L_{\nabla N_{m}}^2 + L_{N_{m}}^2 \ell_{m+1}^2},
	\end{align*}
\else
	\begin{equation*}
	L_N = \sqrt{D_u^2 L_{N_{m}}^2 + B_{N_{m}}^2 + 1}, \quad
	L_{\nabla N} = \sqrt{3 L_{N_{m}}^2 \ell_{m+1} + \ell_{m+1}^2  D_u^2 L_{\nabla N_{m}}^2 + L_{N_{m}}^2 \ell_{m+1}^2},
	\end{equation*}
\fi
\end{theorem}
In the corollary below, we solve the recursive formulas to get simpler (but less tight) expressions of the upper bounds of the constants.

\begin{cor}\label{cor:Lip-consts-iteration-solved}
Let $\ell := \max_{1 \leq u \leq m}\{ \ell_u \}$. The iteratively defined constants of Theorem \ref{rem:lipschitz of NN} can be upper bounded for $1 \leq u \leq m$ by
\if\twocol1
	\begin{align*}
	L_{N_u}^2 &\leq (\ell^2 B_{\Omega})^{2(u-1)} (\sigma'_{\max})^{2 u} (S^2+1)\\
		& \quad +  \sum_{k=1}^{u-1} (\ell^2 B_{\Omega})^{2(k-1)} (\sigma'_{\max})^{2 k} (\ell \sigma_{\max}^2 + 1), \\
	L_{\nabla N_u}^2 &= \mathcal{O}\Big(  u (\sigma_{\max}'')^2 \sigma_{\max}^4 2^{(u-1)} \ell^{(10u-9)} \\
		&\qquad \quad \cdot \left( \sigma_{\max}' B_{\Omega} \right)^{4(u-1)} (S^4+1) \Big).
	\end{align*}
\else
	\begin{align*}
	L_{N_u}^2 &\leq (\ell^2 B_{\Omega})^{2(u-1)} (\sigma'_{\max})^{2 u} (S^2+1) 
		+  \sum_{k=1}^{u-1} (\ell^2 B_{\Omega})^{2(k-1)} (\sigma'_{\max})^{2 k} (\ell \sigma_{\max}^2 + 1), \\
	L_{\nabla N_u}^2 &= \mathcal{O}\Big(  u (\sigma_{\max}'')^2 \sigma_{\max}^4 2^{(u-1)} \ell^{(10u-9)} \, \left( \sigma_{\max}' B_{\Omega} \right)^{4(u-1)} (S^4+1) \Big).
	\end{align*}
\fi
\end{cor}

Now we derive upper bounds on the Lipschitz constants of the objective function $\Phi$ and its gradient.
\begin{theorem}\label{rem:lipschitz of loss}\label{thm:lipschitz of loss}
We assume that the space of parameters $\Omega$ is non-empty, open and bounded in the maximum norm by $0<B_{\Omega}< \infty$. 
Let $Z \sim \P$ be a random variable following the distribution of the training samples and assume that $S:= \lVert \operatorname{proj}_x(Z) \rVert$ is a random variable in $L^2(\P)$, i.e. $\E[S^2] < \infty$. Here $\lVert \cdot \rVert$ denotes the norm \eqref{eq:Euclidean-like norm-Intro} and $\zeta_x$ of Theorem \ref{thm:lipschitz of NN} as replaced by $S$.
Then, the objective function $\Phi$ and its gradient $\nabla \Phi$ are Lipschitz continuous with constants $L_{\Phi}$ and $L_{\nabla \Phi}$. This also implies that $\nabla \Phi$ is uniformly bounded by $B_{\nabla \Phi} = L_\Phi$. Using the constants of Theorem \ref{thm:lipschitz of NN} we define the random variables
\if\twocol1
	\begin{align*}
	L_{\phi} & = B_{\nabla \phi}  =  g'_{\max} \sqrt{D_{m+1}^2 L_{N_{m}}^2 + B_{N_{m}}^2 + 1}, \\
	L_{\nabla \phi} &= \sqrt{\alpha_{m+1} + \beta_{m+1}},\\
	\alpha_{m+1} &= \max\lbrace 3 L_{N_{m}}^2  ((g'_{\max})^2 + (g''_{\max})^2 D_{m+1}^2 B_{N_{m}}^2) \\
			& \quad \quad \quad \quad \quad + 2 (g''_{\max})^2 D_{m+1}^2 L_{N_{m}}^2 \, , \\ 
			& \quad \quad \quad \; \; (g''_{\max})^2 ( B_{N_{m}}^2 + 1 )(3 B_{N_{m}}^2 + 2 ) \rbrace,  \\
	\beta_{m+1} &= ( g'_{\max} D_{m+1} L_{\nabla N_{m}} +  g''_{\max} D_{m+1}^2  L_{N_{m}}^2)^2 \\
			& \quad + L_{N_{m}} \big( g'_{\max} + D_{m+1} g''_{\max} (B_{N_{m}}^2 + 1)^{1/2} \big)^2,
	\end{align*}
\else
	\begin{align*}
	L_{\phi} &  =  g'_{\max} \sqrt{D_{m+1}^2 L_{N_{m}}^2 + B_{N_{m}}^2 + 1}, \quad
	L_{\nabla \phi} = \sqrt{\alpha_{m+1} + \beta_{m+1}},\\
	\alpha_{m+1} &= \max\lbrace 3 L_{N_{m}}^2  ((g'_{\max})^2 + (g''_{\max})^2 D_{m+1}^2 B_{N_{m}}^2) + 2 (g''_{\max})^2 D_{m+1}^2 L_{N_{m}}^2 \, , \\ 
			& \quad \quad \quad \; \; (g''_{\max})^2 ( B_{N_{m}}^2 + 1 )(3 B_{N_{m}}^2 + 2 ) \rbrace,  \\
	\beta_{m+1} &= ( g'_{\max} D_{m+1} L_{\nabla N_{m}} +  g''_{\max} D_{m+1}^2  L_{N_{m}}^2)^2 + L_{N_{m}} \big( g'_{\max} + D_{m+1} g''_{\max} (B_{N_{m}}^2 + 1)^{1/2} \big)^2,
	\end{align*}
\fi
and get the following estimates for the above defined constants:
\begin{equation*}
L_{\Phi} = B_{\nabla \Phi}= \E[L_{\phi}], \quad L_{\nabla \Phi} = \E[L_{\nabla \phi}].
\end{equation*}
\end{theorem}
%
%
Note that since $S$ is now a random variable rather than a constant,  $L_{\phi}, L_{\nabla \phi}, B_{\nabla \phi}$, all depending on $S$  through the dependence on $L_{N_1}$ and  $L_{\nabla N_1}$, therefore are random variables as well.
%
%

\section{Discussion of Theorems}\label{sec:Discussion of Theorems}

\subsection{Local bounds}\label{sec:Local bounds}

Since we assume that the network parameters $\Theta$ lie in the bounded set $\Omega$, we compute local rather than global bounds on the Lipschitz constants.

\textbf{Choice of norm.}
We chose to use the maximum norm instead of the 2-norm for the bound of the network parameters, because with the 2-norm one has to consider how the weights are distributed across the layers, which leads to a complex optimization problem (cf Appendix \ref{sec: Bounding Omega with 2-norm instead of infty-norm}). However, since we consider finite dimensional spaces, both norms are equivalent. 

\textbf{Natural assumption.} 
The assumption that $\Omega$ is bounded is natural. 
For the majority of problems where neural networks can be applied successfully, the loss function numerically converge to some stationary point. Therefore, the parameters can be chosen to only take values in a bounded region. \cite{baes2019lowrank} empirically confirmed this for their optimization problem in Figure 4.  

\textbf{Enforcing bounds on the parameters.}
In several use-cases the network weights are bounded by hard-coded projections or clippings, as for example in the original Wasserstein-GAN \citep{arjovsky2017wasserstein}. This commonly used method guarantees that our assumption on $\Omega$ is satisfied.\\
Moreover, regularisation techniques as for example $L^2$-regularisation outside a certain domain, can be used to essentially guarantee bounds on the 2-norm of the parameters, hence implying our assumption on $\Omega$. A similar approach was used for example in \cite{Ge:2016}.

\textbf{Unbounded biases.}
The assumption on the boundedness of $\Omega$ can be slightly weakened. In particular, it is enough to assume that $\lVert (A^{(1)}, \dotsc, A^{(m+1)}) \rVert_{\infty}$ is bounded, while the biases can be arbitrary. 
Indeed, revisiting the proof of Lemma \ref{lem:induction-step}, we see that only $\lVert  A^{(u)}\rVert_{\infty}$ needs to be bounded by $\tilde{D}$, hence we could replace our assumption that $\lVert \Theta_{m+1} \rVert_{\infty} < B_{\Omega}$ by the  weaker assumption that $\lVert (A^{(1)}, \dotsc, A^{(m+1)}) \rVert_{\infty} < B_{\Omega}$.

\subsection{Global bounds cannot exist}
For upper bounds on the Lipschitz constants to exist, the weight matrices have to be bounded.
This becomes obvious from the following counterexample.

\begin{example} \label{exa:Lip const not improvable}
We assume to have a $1$-dimensional input $\zeta_x = 0$ and a $1$-dimensional output and use $m \in \N$ layers each with $\ell$ hidden units. Furthermore we use as activation function a smoothed version of
\[
x \mapsto \sigma(x):= -c \, 1_{\{x \leq -1\}} + c x \,1_{\{-1 < x < 1\}} + c  \, 1_{\{x \geq 1\}},
\]
coinciding on $(-1,1)$, for some $c >0$ such that $\sigma'_{\max} = c$.
Let us define $\alpha^u_{i,j} := A^{(u)}_{i,j}$, $\beta^u_{i} := b^{(u)}_i$ and $\alpha := (\alpha_{i,j}^u)_{i,j,u}$ and $\beta := (\beta_{i}^u)_{i,u}$, where $i,j,u$ range over all indices for which the $\alpha_{i,j}^u$ and $\beta_{i}^u$ are defined.
If the input to each layer lies in $(-1,1)$, which we can guarantee by choosing the biases small enough, since the input is fixed to be $0$,
we can write the neural network output as
$$
N_{m+1}(\Theta) = \beta_1^{m+1} + c\sum_{i=1}^\ell \alpha^{m+1}_{1,i}  x_i^{m},
$$
where for  $1 \leq u \leq m$
\begin{equation*}
x_i^{u}  :=  \beta_i^{mu} + c\sum_{j=1}^\ell \alpha^{u}_{i,j}  x_j^{u-1}, \quad x_i^1 = \beta^1_i.
\end{equation*}
Choosing all $\alpha_{i,j}^u := \gamma > 1$ and $\beta_i^1 = \delta$, $\beta_i^u = 0$ for $u > 1$ where $\delta < (c \gamma)^m \ell^{m-1}$ (to ensure that the output of each layer is bounded by $1$), we have for $\Theta := (\alpha, \beta)$
\begin{equation*}
N_{m+1}(\Theta) = c \sum_{\nu_{m+1}=1}^\ell \gamma \dotsb c \sum_{\nu_2=1}^\ell \gamma \delta = (c \gamma \ell)^m \delta.
\end{equation*}
Choosing $\tilde{\Theta} := (\alpha, -\beta)$, we similarly have $N_{m+1}(\tilde\Theta) = -(c \gamma \ell)^m \delta$. Taking the differences,
\begin{equation*}
\lVert \Theta - \tilde{\Theta} \rVert = (2 \delta) \sqrt{\ell} ,
\end{equation*}
\begin{equation*}
\lVert N_{m+1}(\Theta) - N_{m+1}(\tilde\Theta) \rVert = 2 (c \gamma \ell)^m \delta,
\end{equation*}
we get a lower bound for the Lipschitz constant
\begin{equation*}
L_{N_m} \geq \frac{\lVert N_{m+1}(\Theta) - N_{m+1}(\tilde\Theta) \rVert}{\lVert \Theta - \tilde{\Theta} \rVert} = (\sigma^\prime_{\max} \gamma )^m \ell^{m-1/2}
\end{equation*}
Hence, if only one of the $\alpha_{i,j}^u$, $2 \leq u \leq m$, is unbounded, there can not exist an upper bound for the Lipschitz constant $L_{N_m}$.
\end{example}

Similar examples can be found, showing that also $\alpha_1$ needs to be bounded.
In particular, the assumption that $\Omega$ is bounded cannot be weakened further than discussed before.

Moreover, this example also shows that it is not enough to assume that the weight matrices come from an unbounded distribution with finite moments, as e.g. a Gaussian distribution, since then for any $k \in \N$ we have that $\P(L_{N_m} > k) > 0$.

\subsection{Activation functions}\label{sec:Activation functions}

We made the assumption that the activation functions $\sigma_u$ are twice differentiable and bounded. This includes the classical sigmoid and $\tanh$ functions, but excludes the often used ReLU function $x \mapsto \max\{ 0, x\}$. However, Theorem \ref{rem:lipschitz of NN} can be extended to slight modifications of ReLU, which are made twice differentiable by smoothing the kink at $0$. Indeed, if $\sigma''_{\max}$ exists, the only part of the proof that has to be adjusted is the computation of $B_{N_u}$.
Since ReLU is either the identity or $0$, the norm of its output is bounded by the norm of its input, which yields $B_{N_1} = D_1 S + \sqrt{\ell_1} B_{\Omega}$ and $B_{N_u} = D_u B_{N_{u-1}} + \sqrt{\ell_u} B_{\Omega}$. To take account for the smoothing of the kink, a small constant $\varepsilon$ can be added, which equals the maximum difference between the smoothed and the original version of ReLU.

One can not hope for more. In particular, if the activation function has a kink, meaning that the first derivative is discontinuous, the derivative of the NN is also discontinuous, i.e. not Lipschitz continuous.

\subsection{Beyond Fully Connected Layers}
While we  consider feed-forward neural networks with fully connected layers, it is easy to extend our results to convolutional layers, since they can also be represented as fully connected layers. Therefore our results can be applied right away. Since many of the weights of the fully connected representation of a convolutional layer are $0$, this can be taken into account by adjusting $D_u$, the bound on the norm of the weights, accordingly.
For other types of layers, similar consideration might lead to extensions of our results. However, we leave a  thorough study of this to future work.

\subsection{Gap between upper and lower bounds.}

\textbf{Bounds for $L_{N}$.}
Example \ref{exa:Lip const not improvable} together with the (rather crude) upper bounds in Corollary \ref{cor:Lip-consts-iteration-solved} imply that
\begin{equation*}
\begin{split}
(\sigma^\prime_{\max} & B_\Omega )^m  \ell^{m-1/2} 
\leq L_{N}  \\
& \leq   ( \sigma'_{\max} B_{\Omega})^{m} \ell^{2m} \sqrt{ m (S^2+1) (\ell \sigma_{\max}^2 + 1)}.
\end{split}
\end{equation*}

\textbf{Bounds for $L_{\nabla N}$.} 
The following Example \ref{ex:Counterexample for lower bound} together with Corollary \ref{cor:Lip-consts-iteration-solved} imply that
\begin{equation*}
\begin{split}
(&\sigma^\prime_{\max})^m  ( B_\Omega   \ell)^{m-1} 
\leq L_{\nabla N}  \\
& \leq   c \,m \,\sigma_{\max}'' \sigma_{\max}^2 2^{m/2} \ell^{5(m-1)} \left( \sigma_{\max}' B_{\Omega} \right)^{2m} \sqrt{(S^4+1)},
\end{split}
\end{equation*}
where $c$ is a constant.

In particular, both Lipschitz constants grow exponentially in the number of layers, and for a fixed number of layers polynomially in $\sigma^\prime_{\max}, B_\Omega, \ell$.

\begin{example} \label{ex:Counterexample for lower bound}
We use the same architecture and notation as in Example \ref{exa:Lip const not improvable}.
Then we get that
\begin{equation*}
\begin{split}
\frac{\partial N_{m}(\Theta) }{\partial_{\beta^u_i}} = \Big( & c \sum^{\ell}_{i_1=1} \alpha_{1, i_1}^{m+1} \dotsb  \\
& \dotsb c \sum^{\ell}_{i_{m+1-u}=1} \alpha^{u+2}_{i_{m-u}, i_{m+1-u}} \Big)  \alpha^{u+1}_{i_{m+1-u}, i}\,.
\end{split}
\end{equation*}
Setting $\alpha_{i,j}^{u} := \gamma$ and $\beta_i^u := 0$, we therefore have for $\Theta:=(\alpha, \beta)$ that
$\partial N_{m}(\Theta)/\partial_{\beta_i}  =  (c \gamma)^{m+1-u} \ell^{m-u}$. 
Moreover, choosing $\tilde\Theta:=(\tilde\alpha, \beta)$ with $\tilde\alpha_{1,i}^{m+1} := -\gamma$ and $\tilde\alpha_{1,i}^{u} := \gamma$ for $u<m+1$, we have $\partial N_{m}(\tilde\Theta)/\partial_{\beta_i}  =  -(c \gamma)^{m+1-u} \ell^{m-u}$.
Then the differences are
\begin{equation*}
\lVert \Theta - \tilde\Theta \rVert^2= \lVert \alpha - \tilde\alpha \rVert^2  + \lVert \beta - \tilde\beta \rVert^2 = \sum_{i=1}^{\ell} (2\gamma)^2 = \ell (2\gamma)^2,
\end{equation*}
\begin{equation*}
\begin{split}
\lVert \nabla_\beta N_m(\Theta) &- \nabla_\beta N_m(\tilde\Theta) \rVert^2 = \\ 
	&=\sum_{u=1}^{m}\sum_{i=1}^{\ell}\left(2 (c \gamma)^{m+1-u} \ell^{m-u} \right)^2  \\
	&= \sum_{u=1}^{m}(c \gamma \ell)^{2(m+1-u)} \ell^{-1} 2^2 .
\end{split}
\end{equation*}
Therefore, 
\begin{equation*}
\begin{split}
\lVert \nabla_\Theta N_m(\Theta) & - \nabla_\Theta N_m(\tilde\Theta) \rVert^2 = \\
	& = \lVert \nabla_\alpha N_m(\Theta) - \nabla_\alpha N_m(\tilde\Theta) \rVert^2 \\
	&  \; + \lVert \nabla_\beta N_m(\Theta) - \nabla_\beta N_m(\tilde\Theta) \rVert^2 \\
	& \geq (c \gamma \ell)^{2m} \ell^{-1} 2^2 ,
\end{split}
\end{equation*}
hence the Lipschitz constant is lower bounded by
\begin{equation*}
L_{\nabla N_{m+1}} 
	\geq \frac{\lVert \nabla_\Theta N_m(\Theta) - \nabla_\Theta N_m(\tilde\Theta) \rVert}{\lVert \Theta - \tilde\Theta \rVert} 
	 \geq c^{m} (\gamma \ell)^{(m-1)}.
\end{equation*} 
\end{example}

%

\section{Applications}

Assume that the (stochastic) gradient scheme in Algorithm \ref{alg:gradient-method} is applied to minimize the objective function $\Phi$.
\begin{algorithm}[H]
   \caption{Stochastic Gradient descent}
   \label{alg:gradient-method}
\begin{algorithmic}
   \STATE Fix $\Theta^{(1)} \in \Omega, M \in \N$
   \FOR{$j\geq 0$}
   \STATE Sample $\zeta_1, \dotsc, \zeta_M \sim \P$
   \STATE Compute $G_j := \tfrac{1}{M} \sum_{i=1}^M \nabla_{\Theta}{\varphi}(\Theta^{(j)}, \zeta_i)$
   \STATE Determine a step-size  $h_{j}>0$
   \STATE Set $\Theta^{(j+1)}=\Theta^{(j)}-h_{j} G_j$
   \ENDFOR
\end{algorithmic}
\end{algorithm}

In the following examples we use Theorem \ref{rem:lipschitz of loss} to set the step sizes of gradient descent (GD) (Example \ref{cor:convergence of GD}) in the case of a finite training set $\mathcal{Z} = \lbrace \zeta_1, \dotsc, \zeta_N \rbrace$ (in particular using $M = N$) and of stochastic gradient descent (SGD) (Example \ref{cor:convergence of SGD}) with adaptive step-sizes (a state-of-the-art neural network training method first introduced in \cite{Duchi:2011:ASM:1953048.2021068}) in the case of a general training set $\mathcal{Z}$.
In particular, the step sizes respectively hyper-parameters for the step sizes can be chosen depending on the computed estimates for $L_{\nabla \Phi}$, such that the GD respectively SGD method are guaranteed to converge (in expectation). At the same time, these examples give bounds on the convergence rates, which turn out to have an exponential dependence on the number of layers. Example~\ref{cor:convergence of SGD} shows that the result of \cite{Li_Orabona_2019} can be applied in our general NN setting, however, only if gradient clipping (or another regularization method) is used to bound the NN weights. 
\begin{examplex}[Stochastic gradient descent]\label{cor:convergence of SGD}
Assume that the random variable $S:= \lVert \operatorname{proj}_x(Z) \rVert$ lies in $L^2(\P)$, i.e. $\E[S^2] < \infty$.
%
Furthermore, assume that there exists $0<B_{\Omega} < \infty$ such that $\sup_{j \geq 1} \lVert \Theta^{(j)}\rVert_{\infty} < B_{\Omega}$. 
For some $\varepsilon >0$, choose the adaptive step-sizes $h_j$ of the stochastic gradient method in Algorithm \ref{alg:gradient-method} as . 
\begin{equation*}
h_j := \frac{1}{\left( 4L_{\nabla \Phi}^2 + \sum_{i=1}^{j-1} \lVert G_i \rVert^2 + \varepsilon\right)^{\frac{1}{2}}},
\end{equation*}
Then, for every $n \in \N$, 
	\begin{equation*}
	\E\left[\min_{1\leq j \leq n} \| \nabla \Phi (\Theta^{(j)}) \| \right]  \leq \frac{C}{\sqrt{n}},
	\end{equation*}
with the constant $C = \max( 2\gamma, \sqrt{2\gamma}(4L_{\nabla \Phi}^2 + \varepsilon + 4n( L_{\Phi}^2 +  \E[ B_{\nabla\phi}^2]))^{\tfrac{1}{4}} )$ where $\gamma = O\left( \frac{1+\ln n}{1- 2(4L_{\nabla \Phi}^2+\varepsilon)^{-1/2}}\right)$.
In particular, for every tolerance level $\delta>0$ we have
	\begin{equation*}
	n\geq \left( \tfrac{C}{\delta} \right)^{2}  \; \Longrightarrow \; \E\left[\min_{1\leq j \leq n} \| \nabla \Phi (\Theta^{(j)}) \| \right] \leq \delta. \hskip1cm \triangle
	\end{equation*}
\end{examplex}
The statements of Example  \ref{cor:convergence of SGD} are proven in Appendix \ref{sec:Proofs in the ordinary DNN setting} and more details on the constant $C$ are given there as well. These examples are just one possibility how the estimates of Theorem \ref{rem:lipschitz of loss} can be used in practice. Similarly, the step sizes of other gradient descent methods can be chosen and convergence rates can be computed using Theorem \ref{rem:lipschitz of loss}. 
For example \cite{ward2018adagrad} show that the convergence rate of AdaGrad-Norm is in $\mathcal{O}((\ln n)^{1/2} n^{-1/4})$, which is the same rate as in Ex 3.9. Their learning rate scheduler does not depend on $L_{\nabla \Phi}$, but the assumption is needed that $\lVert \nabla \Phi \rVert$ is bounded and the precise convergence rate depends on this upper  bound. Our results can be used to formulate sufficient conditions under which such a bound exists and give the upper bound $L_\Phi$ for it.
Other SGD algorithms with convergence rates $\mathcal{O}(n^{-1/4})$ or better are given in \citep{ghadimi2013stochastic, allenzhu2017natasha,  lei2017nonconvex, fang2018spider, zhou2018stochastic}. There the learning rates always depend on the Lipschitz constant $L_{\nabla \Phi}$.

In \cite{baes2019lowrank}, the same result as presented in Example \ref{cor:convergence of GD} was already used to provide a bound on the convergence rate to a stationary point of their algorithm.

\begin{figure}[htp!]
\centering
\begin{minipage}{0.45\textwidth}
\centering
\includegraphics[width=0.7\textwidth]{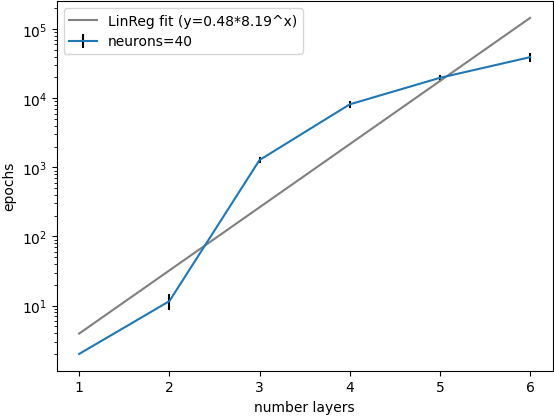}
\caption{Number of epochs needed to achieve a gradient-norm smaller than $\delta = 0.08$ for neural networks with different number of layers. Each hidden layer has $40$ neurons. We observe an exponential growth, as our theoretical studies imply.}
\label{fig:convergence}
\end{minipage}
\end{figure}

\begin{figure}[htp!]
\centering
\begin{minipage}{0.45\textwidth}
\centering
\includegraphics[width=0.8\textwidth]{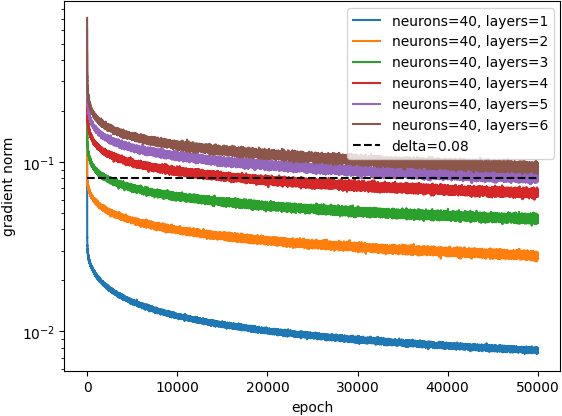}
\caption{Norms of gradients of loss function during training, for neural networks with different number of layers.} 
\label{fig:gradients}
\end{minipage}

\end{figure}

\textbf{Empirical analysis of convergence.}
In order to confirm our theoretical analysis on the bounds, we train feed-forward neural networks of different sizes on the classification task of the MNIST dataset \citep{lecun-mnisthandwrittendigit-2010}.
All layers of the networks have $40$ hidden neurons and the AdaGrad optimizer \cite{ward2018adagrad}  with fixed learning rate $0.01$  is used to train the networks for $50'000$ epochs.
Weight-clipping is used to enforce an $\infty$-norm smaller $1$ of all the weights. During training a dropout rate of $0.1$ and a mini-batch size of $500$ are used.
In Figure~\ref{fig:convergence}, we fix $\delta = 0.08$ and report the number of epochs needed for the neural networks of different depths to achieve a gradient smaller than $\delta$.  We fit a linear regression model to the logarithms of the mean values, reassuring that the growth is exponential in the number of layers. Moreover, we show the evolution of the norms of the gradients of the loss function (on the training set) together with $\delta$ in Figure~\ref{fig:gradients}. 

\section{Deep neural networks as controlled ODEs}\label{sec:Deep neural networks as controlled ODEs}
\subsection{Framework \& definitions}\label{sec:Framework definitions}
We introduce a slightly different notation than in the previous section. 
Let $n \in \N$ and $\Omega \subset \R^n$. Let $\ell \in \N$ be the fixed dimension of the problem. In particular, if we want to define a neural network $\mathcal{N}$ mapping some input $x$ of dimension $\ell_0 \in \N$ to an output $\mathcal{N}(x)$ of dimension $\ell_2 \in \N$ with $\ell_1 \in \N$ the maximal dimension of some hidden layer $\tilde{\mathcal{N}}(x)$, then we set $\ell := \max\lbrace \ell_0, \ell_1, \ell_2 \rbrace$. 
We use ``zero-embeddings'' to write (by abuse of notation) $x, \mathcal{N}(x), \tilde{\mathcal{N}}(x) \in \R^{\ell}$, i.e. we identify $x \in \R^{\ell_0}$ with $(x^\top,0)^\top \in \R^{\ell}$.
This is important, since we want to describe the evolution of an input through a neural network to an output by an ODE, which means that the dimension has to be fixed and cannot change. To do so, we fix $d \in \N$ and define for $1 \leq i \leq d$ vector fields 
\begin{equation*}
V_i : \Omega \times \R_{\geq 0} \times \R^{\ell} \to \R^{\ell}, (\theta, t, x) \mapsto V_i^{\theta} (t, x),
\end{equation*}
which are \emph{c\`agl\`ad} in the second variable and Lipschitz continuous in $x$.
Furthermore, we define  scalar \emph{c\`adl\`ag} functions for $1 \leq i \leq d$, which we refer to as \emph{controls}
\begin{equation*}
u_{i}: \R_{\geq 0} \to \R, t \mapsto u_i(t),
\end{equation*}
which are assumed to have finite variation (also called bounded variation) and start at 0, i.e. $u_i(0)=0$. With these ingredients, we can define the following \emph{controlled ordinary differential equation} (controlled ODE)
\begin{equation}\label{eq:controlled ODE}
dX_{t}^{\theta} = \sum_{i=1}^d V_{i}^{\theta}\big(t,X_{t-}^{\theta}\big)du_{i}(t), \; X_0^{\theta} = x,
\end{equation}
where $x \in \R^{\ell}$ is the starting point, respectively input to the ``neural network''. We fix some $T > 0$. $X^{\theta, x}$ is called a solution of \eqref{eq:controlled ODE}, if it satisfies for all $0 \leq t \leq T$,
\begin{equation}\label{eq:solved controlled ODE}
X_{t}^{\theta,x} = x + \sum_{i=1}^d \int_{0}^t V_{i}^{\theta}\big(s,X_{s-}^{\theta,x}\big)du_{i}(s).
\end{equation}
Then \eqref{eq:solved controlled ODE} describes the evolution of the input $x$ through a ``neural network'' to the output $X_T^{\theta,x}$. Here, the ``neural network'' is defined by $V_i^{\theta}$ and $u_{i}$ for $1 \leq i \leq d$.
\begin{rem}
The assumption on $u_i$ to have finite variation is needed for the integral \eqref{eq:solved controlled ODE} to be well defined. 
Indeed, a deterministic c\`adl\`ag function of finite variation is a special case of a semimartingale, whence we could also take $u_i$ to be semimartingales.
\end{rem}
Before we discuss this framework, we define the loss functions similarly to Section \ref{sec:Problem set-up}. Let $\mathcal{Z} \subset \R^{\ell} \times \R^{k}$ be the set of ($0$-embedded) training samples, again equipped with a sigma algebra and a probability measure $\P$. Let $Z \sim \P$ be a random variable. For a fixed function $g : \R^{\ell} \times \R^{k} \to \R, \; (x,y) \mapsto g(x,y)$, we define the \emph{loss} (or \emph{objective} or \emph{cost}) \emph{function} by
\if\twocol1
	\begin{align*}
	\varphi &: \Omega \times \mathcal{Z} \to \R,  & (\theta, (x,y)) &\mapsto g(X_T^{\theta, x}, y),\\
	\Phi &: \Omega \to \R,  & \theta &\mapsto \E[\varphi(\theta, Z)].
	\end{align*}
\else
	\begin{align*}
	\varphi : \Omega \times \mathcal{Z} \to \R,  \; (\theta, (x,y)) \mapsto g(X_T^{\theta, x}, y),\quad\text{and}\quad
	\Phi : \Omega \to \R,  \; \theta \mapsto \E[\varphi(\theta, Z)].
	\end{align*}
\fi
The framework \eqref{eq:solved controlled ODE} is much more general and powerful than the standard neural network definition. 
In Example \ref{exa:CODE-NN equivalence} we show that the neural network $\mathcal{N}_{\Theta_{m+1}}$ defined in \eqref{eq:NN definition} is a special case of the controlled ODE solution \eqref{eq:solved controlled ODE}.
This example clarifies why we speak of a solution $X^{\theta, x}$ of \eqref{eq:solved controlled ODE} as a ``neural network'', respectively the evolution of the input $x$ through a neural network. If $u$ respectively $u_i$ are not pure step functions, \eqref{eq:solved controlled ODE} defines a neural network of ``infinite depth'', which we refer to as \emph{continuously deep neural networks}. Their output can be approximated using a stepwise scheme to solve ODEs. Doing this, the continuously deep neural network is approximated by a deep neural network of finite depth. 
Using modern ODE solvers with adaptive step sizes as proposed in \cite{Chen2018}, the depth of the approximation and the step sizes change depending on the wanted accuracy and the input.

\subsection{Gradient and existence of solutions }
\label{sec: Gradient and existence of solutions}

Although we are in a deterministic setting, it is reasonable to make use of \emph{It\^o calculus} (also called stochastic calculus) in the above framework, since integrands are predictable. 
See for instance \cite{Pro1992} for an extensive introduction. 
We make use of the typical \emph{differential notation} that is common in stochastic calculus (as for example in \eqref{eq:controlled ODE}) and we treat our ODEs with methods for \emph{stochastic differential equations} (SDEs). Again we emphasize that all $u_i$ could be general semimartingales.

First we note that by Theorem 7 of Chapter V in \cite{Pro1992}, a solution of \eqref{eq:solved controlled ODE} exists and is unique, given that all $V_i^{\theta}(t,x)$ are Lipschitz continuous in $x$.
Starting from \eqref{eq:solved controlled ODE}, we derive the ODE describing the first derivative of $X^{\theta}_t$ with respect to $\theta$. For this, let us define $\partial X_{t}^{\theta} :=\frac{\partial X_{t}^{\theta}}{\partial \theta}$,  $\partial V_{i}^{\theta} :=\frac{ \partial V_{i}^{\theta}}{\partial \theta}$ and for $a,b \in \lbrace x, \theta \rbrace$ we use the standard notation $\partial_{a}V_{i}^{\theta} := \frac{ \partial }{\partial a} V_{i}^{\theta}(t,x)$ and $\partial_{a b}V_{i}^{\theta} := \frac{ \partial }{\partial a} \frac{ \partial }{\partial b} V_{i}^{\theta}(t,x)$.
Assuming that all required derivatives of $V_i^{\theta}$ exist, we have
\if\twocol1
	\begin{equation*}
	\begin{split}
	\partial X_t^{\theta} = \frac{\partial X_{t}^{\theta}}{\partial \theta} &= \sum_{i=1}^d \int_{0}^t \frac{\partial}{\partial \theta} \left(V_{i}^{\theta}\left(s, X_{s-}^{\theta}\right)\right) du_{i}(s)\\
		&= \int_{0}^t \sum_{i=1}^d \left(\partial V_{i}^{\theta}\left(s, X_{s-}^{\theta}\right)\right.\\ 
		 &\qquad \qquad  + \left.   \partial_x V_{i}^{\theta}\left(s,X_{s-}^{\theta}\right) \partial X_{s-}^{\theta}\right)du_{i}(s)\,.
	\end{split}
	\end{equation*}
\else
	\begin{equation*}
	\begin{split}
	\partial X_t^{\theta} = \frac{\partial X_{t}^{\theta}}{\partial \theta} &= \sum_{i=1}^d \int_{0}^t \frac{\partial}{\partial \theta} \left(V_{i}^{\theta}\left(s, X_{s-}^{\theta}\right)\right) du_{i}(s)\\
		&= \int_{0}^t \sum_{i=1}^d \left(\partial V_{i}^{\theta}\left(s, X_{s-}^{\theta}\right) 
		+ \partial_x V_{i}^{\theta}\left(s,X_{s-}^{\theta}\right) \partial X_{s-}^{\theta}\right)du_{i}(s)\,.
	\end{split}
	\end{equation*}
\fi
Therefore, we obtain the following controlled ODE (with differential notation)
\if\twocol1
	\begin{equation}
	\label{ODE}
	\begin{split}
	d\partial X_{t}^{\theta} &= \displaystyle \sum_{i=1}^d \left(\partial V_{i}^{\theta}\left(t, X_{t-}^{\theta}\right) \right.\\ 
	      &\qquad \quad  + \left. \partial_x V_{i}^{\theta}\left(t,X_{t-}^{\theta}\right) \partial X_{t-}^{\theta}\right) du_{i}(t), \\
	      \partial X_{0}^{\theta} &= 0 \in \mathbb{R}^{\ell \times n}.
	\end{split}
	\end{equation}
\else
	\begin{equation}
	\label{ODE}
	\begin{split}
	d\partial X_{t}^{\theta} &= \displaystyle \sum_{i=1}^d \left(\partial V_{i}^{\theta}\left(t, X_{t-}^{\theta}\right)  +  \partial_x V_{i}^{\theta}\left(t,X_{t-}^{\theta}\right) \partial X_{t-}^{\theta}\right) du_{i}(t), \\
	      \partial X_{0}^{\theta} &= 0 \in \mathbb{R}^{\ell \times n}.
	\end{split}
	\end{equation}
\fi
We remark that \eqref{ODE} is a linear ODE, and therefore, by Theorem 7 of Chapter V in \cite{Pro1992}, a unique solution exists, given that all $\partial V_i$ and $\partial_x V_i$ are uniformly bounded.

\subsection{Lipschitz regularity in the setting of controlled ODE}\label{sec:Lipschitz regularity in the setting of controlled ODE}
In the following, we provide similar results for the controlled ODE setting as for the standard DNN setting in Section \ref{sec:Ordinary deep neural network setting}. 
The proofs are again given in Appendix \ref{sec:Proofs in the controlled ODE setting}.

Let us denote the \emph{total variation process} (cf. Chapter I.7 \cite{Pro1992}) of $u_i$ as $\lvert u_i \rvert$. We then define $\upsilon := \sum_{i=1}^d \lvert u_i \rvert$,
which is an increasing function of finite variation with $\upsilon(0) = 0$. Furthermore, we define $B_\upsilon := \upsilon(T)$ and note that $\sum_{i=1}^d \int_0^T d \lvert  u_i \rvert = B_\upsilon$.

With this we are ready to state our main results of this section. 
We start with the equivalent result to Theorem \ref{thm:lipschitz of NN}, giving bounds on the Lipschitz constants of the neural network and its gradient.

\begin{theorem}\label{thm:Lipschitz continuity of X}
Let $\Omega$ be non-empty and open.
We assume that there exist constants
$B_V, B_{\partial_{\theta} V}, B_{\partial_{\theta \theta} V}, B_{\partial_{x \theta} V}, B_{\partial_{\theta x} V}, B_{\partial_{x x} V} \geq 0$ and $p_{\theta}, p_{\theta \theta}, p_{x \theta}, p_{\theta x}, p_{x x} \in \R$
such that for all $1 \leq i \leq d$, $\theta \in \Omega$, $0 \leq t \leq T$ and $x \in \R^{\ell}$ we have 
\if\twocol1
\begin{subequations}\label{eq:assump thm Lipschitz continuity of X}
	\begin{align}
	\lVert V_i^{\theta}(t, x) \rVert &\leq B_V (1 +\lVert x \rVert), \\
	\lVert \partial_{\theta} V_i^{\theta}(t, x) \rVert &\leq B_{\partial_{\theta} V} (1 + \lVert x \rVert^{p_{\theta}}),
	\end{align}
\end{subequations}
\else
\begin{equation}\label{eq:assump thm Lipschitz continuity of X}
	\lVert V_i^{\theta}(t, x) \rVert \leq B_V (1 +\lVert x \rVert), \quad\text{and}\quad
	\lVert \partial_{\theta} V_i^{\theta}(t, x) \rVert \leq B_{\partial_{\theta} V} (1 + \lVert x \rVert^{p_{\theta}}),
\end{equation}
\fi

and similarly for $\partial_{\theta \theta} V_i$, $\partial_{x \theta} V_i$,  $\partial_{\theta x} V_i$ and  $\partial_{x x} V_i$.
We also assume that for any $0 \leq i \leq d$, $0\leq t \leq T$ and $\theta \in \Omega$ 
the map 
\begin{equation*}
\R^{\ell} \to \R^{\ell}, \quad x \mapsto  V_i^{\theta}(t,x)
\end{equation*}
is Lipschitz continuous with constants $L_{V_x}$ independent of $i,t$ and $\theta$. 
Then, for any fixed training sample $(x,y) \in \mathcal{Z}$,
the neural network output $X_T^{\theta, x}$ is uniformly bounded in $\Omega$ by a constant $B_{X}$ and the map and its gradient 
\if\twocol1
	\begin{equation*}
	\begin{split}
	&\Omega \to \R^{\ell}, \quad \theta \mapsto X_T^{\theta, x}, \\
	&\Omega \to \R^{\ell \times n}, \quad \theta \mapsto \partial_{\theta} X_T^{\theta, x},
	\end{split}
	\end{equation*}
\else
	\begin{equation*}
	\Omega \to \R^{\ell}, \; \theta \mapsto X_T^{\theta, x}, \quad\text{and}\quad
	\Omega \to \R^{\ell \times n}, \; \theta \mapsto \partial_{\theta} X_T^{\theta, x},
	\end{equation*}
\fi
are Lipschitz continuous on $\Omega$ with constants $L_{X}$ and $L_{\partial X}$. This also implies that $\partial_{\theta} X_T^{\theta, x}$ is uniformly bounded by $B_{\partial X} = L_X$. Upper bounds for these constants can be computed as
\begin{align*}
B_X &= (\lVert x \rVert + B_V B_\upsilon) \exp(B_V B_\upsilon), \\
L_X &= B_{\partial_{\theta} V} (1+B_X^{p_{\theta}}) B_\upsilon \exp(L_{V_x} B_\upsilon), \\
C_{\theta \theta} &= B_\upsilon \big[ B_{\partial_{\theta \theta} V} (1+B_X^{p_{\theta \theta}}) +  B_{\partial_{x \theta} V} (1+B_X^{p_{x \theta}}) L_X \\
	& + B_{\partial_{\theta x} V} (1+B_X^{p_{\theta x}}) L_X   + B_{\partial_{x x} V} (1+B_X^{p_{x x}}) L_X^2 \big], \\
L_{\partial X} &=C_{\theta \theta} \exp(L_{V_x} B_\upsilon).
\end{align*}
\end{theorem}
A remark about the bounding constants is given in Remark \ref{rem:Schwarz Thm}. Next, we present the equivalent result to Theorem \ref{thm:lipschitz of loss}, giving bounds on the Lipschitz constant of the objective function and its gradient.
\begin{theorem}\label{thm:Lipschitz continuity of Phi - controlled ODE}
We make the same assumptions as in Theorem \ref{thm:Lipschitz continuity of X}.
Furthermore, we assume that for any fixed $y \in \operatorname{proj}_y(\mathcal{Z}) $, the functions 
\if\twocol1
	\begin{align*}
	\R^{\ell} &\to \R, \quad x \mapsto g(x, y), \\
	\R^{\ell} &\to \R^{\ell}, \quad x \mapsto \tfrac{\partial}{\partial x} g(x,y)
	\end{align*}
\else
	\begin{align*}
	\R^{\ell} \to \R, \; x \mapsto g(x, y), \quad\text{and}\quad
	\R^{\ell} \to \R^{\ell}, \; x \mapsto \tfrac{\partial}{\partial x} g(x,y)
	\end{align*}
\fi
are Lipschitz continuous on $\operatorname{proj}_x(\mathcal{Z})$ with constants $L_g, L_{\partial_x g}$ independent of $y$. 
Let $Z\sim \P$ be a random variable following the distribution of the training samples and assume that the random variable $S:= \lVert \operatorname{proj}_x(Z) \rVert$ lies in $L^p(\mathcal{Z}, \mathcal{A}(\mathcal{Z}), \P)$, i.e. $\E[S^p] < \infty$, where $p := \max \lbrace 1, p_{\theta \theta}, p_{x \theta} + p_{\theta}, p_{ \theta x} + p_{\theta}, p_{x x} + 2 p_{\theta} \rbrace$. 
Then, the objective function and its gradient
\if\twocol1
	\begin{equation*}
	\begin{split}
	&\Omega \to \R,  \quad \theta \mapsto \Phi(\theta),\\
	&\Omega \to \R^{n},  \quad \theta \mapsto \nabla\Phi(\theta),
	\end{split}
	\end{equation*}
\else
	\begin{equation*}
	\Omega \to \R,  \; \theta \mapsto \Phi(\theta),\quad\text{and}\quad
	\Omega \to \R^{n},  \; \theta \mapsto \nabla\Phi(\theta),
	\end{equation*}
\fi
are Lipschitz continuous with Lipschitz constants $L_{\Phi}$ and $L_{\nabla \Phi}$. This also implies that $\nabla \Phi$ is uniformly bounded by $B_{\nabla \Phi} = L_{\Phi}$. 
Upper bounds for these constants can be computed as
\if\twocol1
	\begin{align*}
	L_{\Phi} &= \E[L_g L_X], \\
	L_{\nabla \Phi} &= \E[L_{\partial_x g} L_X^2 + L_g L_{\partial X}].
	\end{align*}
\else
	\begin{equation*}
	L_{\Phi} = \E[L_g L_X], \quad L_{\nabla \Phi} = \E[L_{\partial_x g} L_X^2 + L_g L_{\partial X}].
	\end{equation*}
\fi
\end{theorem}

A comparison of the Theorems of this Section with those of Section \ref{sec:Ordinary deep neural network setting} is given in Remark \ref{rem:compareing theorems}.

Theorem \ref{thm:Lipschitz continuity of Phi - controlled ODE} can be used exactly like Theorem \ref{thm:lipschitz of loss} to set the step sizes of gradient descent methods. In particular, if a (stochastic) gradient descent scheme as in Algorithm \ref{alg:gradient-method} is used, we get the same results as in Example \ref{cor:convergence of GD} and \ref{cor:convergence of SGD}.

\section{Conclusion and Discussion}
As  SGD  methods are the most popular for the training of deep neural networks, we analyse the convergence of the loss function of neural networks when trained with SGD methods. One factor that plays an important role in SGD  methods is the Lipschitz constant. 
We studied the Lipschitz constants with respect to the parameters and we provided  upper and lower bounds for the first time. 
We proved that the Lipschitz constants grow exponentially in the number of layers, and for a fixed number of layers polynomially in $\sigma^\prime_{\max}, B_\Omega$ and the number of neurons.
As those bounds are very large for many deep neural network architectures, it is often not reasonable to use them in practice for setting the step sizes of SGD methods. However, they can be used for a theoretical analysis of the convergence rates of SGD methods.
Our goal in this work was to give general Lipschitz bounds and implications of them, that apply to all standard neural network settings. There exist several works analysing one specific case, but we are not aware of any work that gave a general analysis so far. 
Therefore, we believe that this is an useful contribution to the active research on analysing and understanding the convergence behaviour of deep neural networks.

%
%
%
%
%

\if\addackn1
	\section*{Acknowledgement}
	We thank Andrew Allan and Maximilian Nitzschner for proofreading and valuable remarks. 
	Furthermore, we thank Robert A. Crowell and Hanna Wutte for fruitful discussions.
\fi

\if\broaderimpact1
	\section*{Broader Impact}
	Due to the very theoretical nature of this paper, a Broader Impact discussion is not applicable.
\fi

\if\twocol1
	\if\aistats0
		\bibliographystyle{icml2020}
	\else
		\bibliographystyle{icml2020}
	\fi
\else
	\bibliographystyle{unsrtnat}
\fi
\bibliography{Lip}

\begin{thebibliography}{41}
\providecommand{\natexlab}[1]{#1}
\providecommand{\url}[1]{\texttt{#1}}
\expandafter\ifx\csname urlstyle\endcsname\relax
  \providecommand{\doi}[1]{doi: #1}\else
  \providecommand{\doi}{doi: \begingroup \urlstyle{rm}\Url}\fi

\bibitem[Allen-Zhu(2017)]{allenzhu2017natasha}
Allen-Zhu, Z.
\newblock Natasha: Faster non-convex stochastic optimization via strongly
  non-convex parameter, 2017.

\bibitem[Allen-Zhu et~al.(2019)Allen-Zhu, Li, and Song]{allen2019convergence}
Allen-Zhu, Z., Li, Y., and Song, Z.
\newblock A convergence theory for deep learning via over-parameterization.
\newblock In \emph{International Conference on Machine Learning}, pp.\
  242--252. PMLR, 2019.

\bibitem[Arjovsky et~al.(2017)Arjovsky, Chintala, and
  Bottou]{arjovsky2017wasserstein}
Arjovsky, M., Chintala, S., and Bottou, L.
\newblock {Wasserstein GAN}.
\newblock \emph{arXiv preprint arXiv:1701.07875}, 2017.

\bibitem[Arora et~al.(2018)Arora, Ge, Neyshabur, and Zhang]{pmlr-v80-arora18b}
Arora, S., Ge, R., Neyshabur, B., and Zhang, Y.
\newblock Stronger generalization bounds for deep nets via a compression
  approach.
\newblock volume~80 of \emph{Proceedings of Machine Learning Research}, pp.\
  254--263, Stockholmsm{\"a}ssan, Stockholm Sweden, 10--15 Jul 2018. PMLR.
\newblock URL \url{http://proceedings.mlr.press/v80/arora18b.html}.

\bibitem[Baes et~al.(2019)Baes, Herrera, Neufeld, and Ruyssen]{baes2019lowrank}
Baes, M., Herrera, C., Neufeld, A., and Ruyssen, P.
\newblock Low-rank plus sparse decomposition of covariance matrices using
  neural network parametrization, 2019.

\bibitem[Bartlett et~al.(2017)Bartlett, Foster, and
  Telgarsky]{bartlett2017spectrally}
Bartlett, P.~L., Foster, D.~J., and Telgarsky, M.~J.
\newblock Spectrally-normalized margin bounds for neural networks.
\newblock \emph{Advances in neural information processing systems}, 30, 2017.

\bibitem[Brouwer et~al.(2019)Brouwer, Simm, Arany, and
  Moreau]{Brouwer2019GRUODEBayesCM}
Brouwer, E.~D., Simm, J., Arany, A., and Moreau, Y.
\newblock {GRU-ODE-Bayes}: Continuous modeling of sporadically-observed time
  series.
\newblock \emph{NeurIPS}, 2019.

\bibitem[Cao \& Gu(2019)Cao and Gu]{cao2019generalization}
Cao, Y. and Gu, Q.
\newblock Generalization bounds of stochastic gradient descent for wide and
  deep neural networks.
\newblock \emph{Advances in neural information processing systems}, 32, 2019.

\bibitem[Chen et~al.(2018)Chen, Rubanova, Bettencourt, and Duvenaud]{Chen2018}
Chen, T.~Q., Rubanova, Y., Bettencourt, J., and Duvenaud, D.~K.
\newblock Neural ordinary differential equations.
\newblock In \emph{Advances in Neural Information Processing Systems 31}. 2018.

\bibitem[Cohen \& Elliott(2015)Cohen and Elliott]{cohen2015stochastic}
Cohen, S.~N. and Elliott, R.~J.
\newblock \emph{Stochastic calculus and applications}.
\newblock Springer, 2nd edition, 2015.

\bibitem[Combettes \& Pesquet(2019)Combettes and
  Pesquet]{Combettes2019LipschitzCF}
Combettes, P.~L. and Pesquet, J.-C.
\newblock Lipschitz certificates for neural network structures driven by
  averaged activation operators.
\newblock 2019.

\bibitem[Cuchiero et~al.(2019)Cuchiero, Larsson, and
  Teichmann]{cuchiero2019deep}
Cuchiero, C., Larsson, M., and Teichmann, J.
\newblock Deep neural networks, generic universal interpolation, and controlled
  {ODE}s, 2019.
\newblock URL \url{http://arxiv.org/abs/1908.07838}.

\bibitem[Duchi et~al.(2011)Duchi, Hazan, and
  Singer]{Duchi:2011:ASM:1953048.2021068}
Duchi, J., Hazan, E., and Singer, Y.
\newblock Adaptive subgradient methods for online learning and stochastic
  optimization.
\newblock volume~12, pp.\  2121--2159, 2011.

\bibitem[Fang et~al.(2018)Fang, Li, Lin, and Zhang]{fang2018spider}
Fang, C., Li, C.~J., Lin, Z., and Zhang, T.
\newblock Spider: Near-optimal non-convex optimization via stochastic path
  integrated differential estimator, 2018.

\bibitem[Fazlyab et~al.(2019)Fazlyab, Robey, Hassani, Morari, and
  Pappas]{Fazlyab2019EfficientAA}
Fazlyab, M., Robey, A., Hassani, H., Morari, M., and Pappas, G.~J.
\newblock Efficient and accurate estimation of lipschitz constants for deep
  neural networks.
\newblock In \emph{NeurIPS}, 2019.

\bibitem[Ge et~al.(2016)Ge, Lee, and Ma]{Ge:2016}
Ge, R., Lee, J.~D., and Ma, T.
\newblock Matrix completion has no spurious local minimum.
\newblock In \emph{Proceedings of the 30th International Conference on Neural
  Information Processing Systems}, NIPS'16, pp.\  2981--2989, 2016.

\bibitem[Ghadimi \& Lan(2013)Ghadimi and Lan]{ghadimi2013stochastic}
Ghadimi, S. and Lan, G.
\newblock Stochastic first- and zeroth-order methods for nonconvex stochastic
  programming, 2013.

\bibitem[Jia \& Benson(2019)Jia and Benson]{NJSDE}
Jia, J. and Benson, A.~R.
\newblock Neural jump stochastic differential equations.
\newblock 2019.

\bibitem[Jin \& Lavaei(2018)Jin and Lavaei]{Jin2018StabilitycertifiedRL}
Jin, M. and Lavaei, J.
\newblock Stability-certified reinforcement learning: A control-theoretic
  perspective.
\newblock \emph{ArXiv}, abs/1810.11505, 2018.

\bibitem[Kingma \& Ba(2014)Kingma and Ba]{kingma2014adam}
Kingma, D. and Ba, J.
\newblock Adam: A method for stochastic optimization.
\newblock \emph{International Conference on Learning Representations}, 2014.

\bibitem[K{\"o}nigsberger(2013)]{konigsberger2013analysis}
K{\"o}nigsberger, K.
\newblock \emph{Analysis 2}.
\newblock Springer-Verlag, 2013.

\bibitem[Latorre et~al.(2020)Latorre, Rolland, and
  Cevher]{Latorre2020LipschitzCE}
Latorre, F., Rolland, P., and Cevher, V.
\newblock Lipschitz constant estimation for neural networks via sparse
  polynomial optimization.
\newblock In \emph{ICLR 2020}, 2020.

\bibitem[LeCun \& Cortes(2010)LeCun and
  Cortes]{lecun-mnisthandwrittendigit-2010}
LeCun, Y. and Cortes, C.
\newblock {MNIST Handwritten Digit Database}.
\newblock 2010.
\newblock URL \url{http://yann.lecun.com/exdb/mnist/}.

\bibitem[Lei et~al.(2017)Lei, Ju, Chen, and Jordan]{lei2017nonconvex}
Lei, L., Ju, C., Chen, J., and Jordan, M.~I.
\newblock Non-convex finite-sum optimization via scsg methods, 2017.

\bibitem[Li \& Orabona(2019)Li and Orabona]{Li_Orabona_2019}
Li, X. and Orabona, F.
\newblock On the convergence of stochastic gradient descent with adaptive
  stepsizes.
\newblock \emph{Proceedings of Machine Learning Research}, 89, Apr 2019.

\bibitem[Li et~al.(2018)Li, Lu, Wang, Haupt, and Zhao]{li2018tighter}
Li, X., Lu, J., Wang, Z., Haupt, J., and Zhao, T.
\newblock On tighter generalization bound for deep neural networks: Cnns,
  resnets, and beyond.
\newblock \emph{arXiv preprint arXiv:1806.05159}, 2018.

\bibitem[Liu et~al.(2019)Liu, Xiao, Si, Cao, Kumar, and
  Hsieh]{DBLP:journals/corr/abs-1906-02355}
Liu, X., Xiao, T., Si, S., Cao, Q., Kumar, S., and Hsieh, C.
\newblock Neural {SDE:} stabilizing neural {ODE} networks with stochastic
  noise.
\newblock \emph{CoRR}, abs/1906.02355, 2019.
\newblock URL \url{http://arxiv.org/abs/1906.02355}.

\bibitem[Melis et~al.(2017)Melis, Dyer, and Blunsom]{Melis2017OnTS}
Melis, G., Dyer, C., and Blunsom, P.
\newblock On the state of the art of evaluation in neural language models.
\newblock 2017.
\newblock URL \url{http://arxiv.org/abs/1707.05589}.

\bibitem[Nesterov(2013)]{nesterov2013introductory}
Nesterov, Y.
\newblock \emph{Introductory lectures on convex optimization: A basic course},
  volume~87.
\newblock Springer Science \& Business Media, 2013.

\bibitem[Peluchetti \& Favaro(2019)Peluchetti and
  Favaro]{Peluchetti2019InfinitelyDN}
Peluchetti, S. and Favaro, S.
\newblock Infinitely deep neural networks as diffusion processes.
\newblock 2019.

\bibitem[Protter(1992)]{Pro1992}
Protter, P.
\newblock \emph{Stochastic integration and differential equations}.
\newblock Springer-Verlag, 2nd edition, 1992.

\bibitem[Raghunathan et~al.(2018)Raghunathan, Steinhardt, and
  Liang]{Raghunathan2018CertifiedDA}
Raghunathan, A., Steinhardt, J., and Liang, P.
\newblock Certified defenses against adversarial examples.
\newblock \emph{ArXiv}, abs/1801.09344, 2018.

\bibitem[Reddi et~al.(2018)Reddi, Kale, and Kumar]{Reddi2018}
Reddi, S., Kale, S., and Kumar, S.
\newblock On the convergence of adam and beyond.
\newblock In \emph{International Conference on Learning Representations}, 2018.

\bibitem[Rubanova et~al.(2019)Rubanova, Chen, and
  Duvenaud]{DBLP:journals/corr/abs-1907-03907}
Rubanova, Y., Chen, R. T.~Q., and Duvenaud, D.
\newblock Latent odes for irregularly-sampled time series.
\newblock \emph{CoRR}, abs/1907.03907, 2019.
\newblock URL \url{http://arxiv.org/abs/1907.03907}.

\bibitem[Scaman \& Virmaux(2018)Scaman and Virmaux]{Scaman2018LipschitzRO}
Scaman, K. and Virmaux, A.
\newblock Lipschitz regularity of deep neural networks: analysis and efficient
  estimation.
\newblock In \emph{NeurIPS}, 2018.

\bibitem[Tieleman \& Hinton(2012)Tieleman and Hinton]{Tieleman2012}
Tieleman, T. and Hinton, G.
\newblock {Lecture 6.5---RmsProp: Divide the gradient by a running average of
  its recent magnitude}.
\newblock COURSERA: Neural Networks for Machine Learning, 2012.

\bibitem[Tzen \& Raginsky(2019)Tzen and
  Raginsky]{DBLP:journals/corr/abs-1905-09883}
Tzen, B. and Raginsky, M.
\newblock Neural stochastic differential equations: Deep latent gaussian models
  in the diffusion limit.
\newblock \emph{CoRR}, abs/1905.09883, 2019.
\newblock URL \url{http://arxiv.org/abs/1905.09883}.

\bibitem[Ward et~al.(2018)Ward, Wu, and Bottou]{ward2018adagrad}
Ward, R., Wu, X., and Bottou, L.
\newblock Adagrad stepsizes: Sharp convergence over nonconvex landscapes, from
  any initialization, 2018.

\bibitem[Xu et~al.(2015)Xu, Ba, Kiros, Cho, Courville, Salakhudinov, Zemel, and
  Bengio]{pmlr-v37-xuc15}
Xu, K., Ba, J., Kiros, R., Cho, K., Courville, A., Salakhudinov, R., Zemel, R.,
  and Bengio, Y.
\newblock Show, attend and tell: Neural image caption generation with visual
  attention.
\newblock In \emph{Proceedings of the 32nd International Conference on Machine
  Learning}, 2015.

\bibitem[Zhou et~al.(2018)Zhou, Xu, and Gu]{zhou2018stochastic}
Zhou, D., Xu, P., and Gu, Q.
\newblock Stochastic nested variance reduction for nonconvex optimization,
  2018.

\bibitem[Zou et~al.(2018)Zou, Cao, Zhou, and Gu]{zou2018stochastic}
Zou, D., Cao, Y., Zhou, D., and Gu, Q.
\newblock Stochastic gradient descent optimizes over-parameterized deep relu
  networks.
\newblock \emph{arXiv preprint arXiv:1811.08888}, 2018.

\end{thebibliography}

\if\inclapp1
	\if\twocol1
		\clearpage
	\else
		\clearpage
	\fi
	\section*{Appendix}
	\appendix

\section{Auxiliary results in the ordinary DNN setting}\label{sec:Auxiliary results in the ordinary DNN setting}

\begin{example}[Gradient descent]\label{cor:convergence of GD}
Assume that $\mathcal{Z} = \lbrace \zeta_1, \dotsc, \zeta_N \rbrace$ with equal probabilities and that in each step of the gradient method the true gradient of $\Phi$ is computed, i.e. gradient descent and not a stochastic version of it is applied. Furthermore, assume that there exists $0 < B_{\Omega} < \infty$ such that $\sup_{j \geq 1} \lVert \Theta^{(j)}\rVert_{\infty} < B_{\Omega}$. 
Choosing the step sizes $h_j := \tfrac{1}{L_{\nabla\Phi}}$, the following inequality
\begin{equation*}\label{eq:condition-gradient}
\Phi(\Theta^{(j)}) - \Phi(\Theta^{(j+1)}) \geq \tfrac{1}{2 L_{\nabla \Phi}} \lVert \nabla \Phi (\Theta^{(j)}) \rVert^2,
\end{equation*}
is always satisfied as shown in Section~1.2.3 of \cite{nesterov2013introductory}. Furthermore, it follows that for every $n \in \N$ we have  
\begin{equation*}
	\min_{1\leq j \leq n} \| \nabla \Phi (\Theta^{(j)}) \| \leq \tfrac{1}{\sqrt{n}}\Big[2 L_{\nabla \Phi} \big(\Phi(\Theta^{(1)})-\Phi^*\big)\Big]^{1/2}, 
	\end{equation*}
	where $\Phi^*:=\min_{ \lVert\Theta \rVert \leq B_{\Omega}} \Phi(\Theta)$.
	In particular, for every tolerance level $\varepsilon>0$ we have
	\begin{equation*}
	n \geq \tfrac{L_{\nabla \Phi}}{K\varepsilon^2} \big(\Phi(\Theta^{(1)})-\varphi^*\big) \; \Longrightarrow \; \min_{1\leq j \leq n} \| \nabla \Phi (\Theta^{(j)})\| \leq \varepsilon.
	\end{equation*}
\end{example}

\section{Proofs in the ordinary DNN setting}\label{sec:Proofs in the ordinary DNN setting}
Before we start to prove the theorems, we establish some helpful results.	

\begin{lem}\label{lem:induction-step}
Let $n_1, n_2, n_3 \in \N_{>0}$, let $C \in \R^{n_3\times n_2}$, $d \in \R^{n_3}$, $l_2 = n_3(n_2+1)$ and let $\lambda \in \R^{l_2}$ be a flattened version of $(C,d)$. We restrict $\lambda$ to be an element of $\mathcal{L} := \lbrace \lambda \in \R^{l_2} \; | \; \lVert \lambda \rVert_{\infty} < \tilde{D} \rbrace$, for some $\tilde{D}>0$.
Let $\tilde{\psi} : \R \to \R$ be a function with bounded first and second derivatives, i.e. there exist $c_1, c_2 \geq 0$ such that for all $x \in \R$ we have $\lvert \tilde{\psi}^{\prime}(x) \rvert \leq c_1$ and $\lvert \tilde{\psi}^{\prime \prime}(x) \rvert \leq c_2$. Let $\psi: \R^{n_3} \to \R^{n_3}, x \mapsto (\tilde{\psi}(x_1), \dotsc, \tilde{\psi}(x_{n_3}))$ and define $\Psi_{\lambda} : \R^{n_2} \to \R^{n_3}, x \mapsto \psi(C x + d)$. Furthermore, let $l_1 \in \N_{\geq 0}$ and $\kappa \in \mathcal{K} \subset \R^{l_1}$.
Let $\rho_{\kappa} : \R^{n_1} \to \R^{n_2}, x \mapsto \rho_{\kappa}(x)$ be a function depending on the parameters $\kappa$. Let $\mu = (\kappa, \lambda) \in \mathcal{M} := \mathcal{K} \times \mathcal{L} \subset \R^l$, where $l = l_1 + l_2$. For a fixed $\zeta \in \R^{n_1}$ we define the function 
\[\chi : \mathcal{M} \to \R^{n_3}, \mu = (\kappa, \lambda) \mapsto \chi(\mu) := \Psi_{\lambda}(\rho_{\kappa}(\zeta)).\]
We use the notation $\Psi_{\lambda}^{\prime}(x) := \tfrac{\partial}{\partial x} \Psi_{\lambda}(x) $,  $\nabla \Psi_{\lambda}(x) := [\tfrac{\partial}{\partial \lambda_j} (\Psi_{\lambda}(x))_i]_{i,j}$, and similar for $\nabla \rho_{\kappa}(\zeta)$.
If $\kappa \mapsto \rho_{\kappa}(\zeta)$ is Lipschitz continuous with constant $L_1$ and $\kappa \mapsto \nabla \rho_{\kappa}(\zeta)$ with constant $L_2$ and if $\lVert \rho_{\kappa}(\zeta) \rVert \leq B_1$ and $\lVert \nabla \rho_{\kappa}(\zeta) \rVert \leq B_2$, where $0 \leq L_1, L_2, B_1,B_2 < \infty$ and $D := n_2 n_3 \tilde{D}$,
then we have that
\begin{itemize}
\item[i)] $\chi$ is Lipschitz continuous with constant $L_{\chi} = c_1 \sqrt{D^2 L_1^2 + B_1^2 + 1}$,
\item[ii)] $\nabla \chi$ is Lipschitz continuous with constant $L_{\nabla \chi} = \sqrt{m_1 + m_2}$, where
\if\twocol1
	\begin{equation*}
	\begin{split}
	 m_1 &:= \max\lbrace  3  L_1^2 (c_1^2 n_3 + c_2^2 D^2 B_1^2)  \\
	    & \qquad \qquad + 2 c_2^2 D^2 L_1^2,  c_2^2 (B_1^2 + 1) (3 B_1^2 + 2 ) \rbrace,\\
	m_2 &:=  ( n_3 c_1 D L_2  + B_2 c_2 D^2 L_1)^2\\
		& \quad \; +B_2^2 \big( n_3 c_1  + D c_2 (B_1^2 + 1)^{1/2} \big)^2,
	\end{split}
	\end{equation*}
\else
	\begin{equation*}
	\begin{split}
	 m_1 &:= \max\lbrace  3  L_1^2 (c_1^2 n_3 + c_2^2 D^2 B_1^2) 
	 	+ 2 c_2^2 D^2 L_1^2,  c_2^2 (B_1^2 + 1) (3 B_1^2 + 2 ) \rbrace,\\
	m_2 &:=  ( n_3 c_1 D L_2  + B_2 c_2 D^2 L_1)^2
	 +B_2^2 \big( n_3 c_1  + D c_2 (B_1^2 + 1)^{1/2} \big)^2,
	\end{split}
	\end{equation*}
\fi
\item[iii)] the gradient $\nabla \chi (\mu)$ of $\chi$ is bounded by $ B_{\nabla \chi} = L_{\chi}$.
If we also assume that $\tilde{\psi}$ is bounded by $0 < B_3 < \infty$, i.e. for all $x \in \R: \lvert \tilde{\psi}(x) \rvert \leq B_3 $, then $\chi(\mu)$ is bounded by $B_{\chi} = \sqrt{n_3} B_3$.
\end{itemize}
\end{lem}

\begin{proof}[Proof of Lemma \ref{lem:induction-step}]
Let $\mu = (\kappa, \lambda), \bar{\mu} = (\bar{\kappa}, \bar{\lambda}) \in \mathcal{M}$ with $\lambda = (C,d)$ and $\bar{\lambda} = (\bar{C}, \bar{d})$.
For $i)$, we use that $x \mapsto \psi (x)$ is Lipschitz with constant $c_1$ and compute
\if\twocol1
	\begin{equation*}
	\begin{split}
	\lVert  \chi(\mu) &- \chi(\bar{\mu}) \rVert^2 \\
	&= \lVert \psi(C \rho_{\kappa}(\zeta) + d ) - \psi(\bar{C} \rho_{\bar{\kappa}}(\zeta) + \bar{d} ) \rVert^2 \\
		&\leq c_1^2 \big(  \lVert C \rho_{\kappa}(\zeta) + d - C \rho_{\bar{\kappa}}(\zeta) - d \rVert  \\
			& \quad + \lVert C \rho_{\bar{\kappa}}(\zeta) + d  - \bar{C} \rho_{\bar{\kappa}}(\zeta) - \bar{d} \rVert \big)^2 \\
		& \leq  c_1^2 \big( \lVert C \rVert \lVert \rho_{\kappa}(\zeta)  -  \rho_{\bar{\kappa}}(\zeta) \rVert  \\
			 & \quad +\lVert \rho_{\bar{\kappa}}(\zeta) \rVert \lVert C - \bar{C} \rVert + \lVert d - \bar{d} \rVert   \big)^2 \\
		&\leq c_1^2 \big( D L_1 \lVert \kappa - \bar{\kappa} \rVert +	 B_1 \lVert C - \bar{C} \rVert + \lVert d - \bar{d} \rVert  \big)^2 \\
		&\leq c_1^2 (D^2 L_1^2 + B_1^2 + 1) \lVert \mu - \bar{\mu} \rVert^2,
	\end{split}
	\end{equation*}
\else
	\begin{equation*}
	\begin{split}
	\lVert  \chi(\mu) - \chi(\bar{\mu}) \rVert^2
		&= \lVert \psi(C \rho_{\kappa}(\zeta) + d ) - \psi(\bar{C} \rho_{\bar{\kappa}}(\zeta) + \bar{d} ) \rVert^2 \\
		&\leq c_1^2 \big(  \lVert C \rho_{\kappa}(\zeta) + d - C \rho_{\bar{\kappa}}(\zeta) - d \rVert   
		+ \lVert C \rho_{\bar{\kappa}}(\zeta) + d  - \bar{C} \rho_{\bar{\kappa}}(\zeta) - \bar{d} \rVert \big)^2 \\
		& \leq  c_1^2 \big( \lVert C \rVert \lVert \rho_{\kappa}(\zeta)  -  \rho_{\bar{\kappa}}(\zeta) \rVert  
		+\lVert \rho_{\bar{\kappa}}(\zeta) \rVert \lVert C - \bar{C} \rVert + \lVert d - \bar{d} \rVert   \big)^2 \\
		&\leq c_1^2 \big( D L_1 \lVert \kappa - \bar{\kappa} \rVert +	 B_1 \lVert C - \bar{C} \rVert + \lVert d - \bar{d} \rVert  \big)^2 \\
		&\leq c_1^2 (D^2 L_1^2 + B_1^2 + 1) \lVert \mu - \bar{\mu} \rVert^2,
	\end{split}
	\end{equation*}
\fi
where we used the Cauchy--Schwarz inequality in the last step.\\
For $ii)$ we first compute some partial derivatives of the functions under consideration with respect to $C_{i,j}$, $1 \leq i \leq n_3$, $1 \leq j \leq n_2$ and $d_i$, $1 \leq i \leq n_3$ and $\kappa$. Denoting by $e_i$  the canonical basis vectors in $\R^{n_3}$, one has
\begin{subequations}\label{eq:partials-chi}
\begin{align}
\tfrac{\partial}{\partial C_{i,j}} \Psi_{\lambda}(x) &= \tilde{\psi}^{\prime}(C_{i, \cdot} x + d_i) x_j e_i, \\
\tfrac{\partial}{\partial d_i} \Psi_{\lambda}(x) &= \tilde{\psi}^{\prime}(C_{i, \cdot} x + d_i) e_i ,\\
\tfrac{\partial}{\partial \kappa} \chi(\mu) &= \Psi_{\lambda}^{\prime}(\rho_{\kappa}(\zeta)) \, \nabla \rho_{\kappa}(\zeta),\\
\Psi_{\lambda}^{\prime}(x) &= 
	\text{diag}(\psi^{\prime}(C x + d)) C.
\end{align}
\end{subequations}
We then compute the Lipschitz constants of the different partial derivatives of $\chi$. As above, we use the triangle inequality extensively to get:
\if\twocol1
	\begin{equation*}
	\begin{split}
	\lVert & \tfrac{\partial}{\partial C_{i,j}}  \chi(\mu) - \tfrac{\partial}{\partial C_{i,j}} \chi(\bar{\mu}) \rVert^2  \\
		& = \lVert \tfrac{\partial}{\partial C_{i,j}} \Psi_{\lambda}(\rho_{\kappa}(\zeta))  - \tfrac{\partial}{\partial C_{i,j}} \Psi_{\bar{\lambda}}(\rho_{\bar{\kappa}}(\zeta)) \rVert^2 \\
		& \leq \big( \lVert  \tilde{\psi}^{\prime}(C_{i, \cdot} \rho_{\kappa}(\zeta) + d_i)\rVert  \lVert (\rho_{\kappa}(\zeta))_j  - (\rho_{\bar{\kappa}}(\zeta))_j\rVert  \\
		& \quad + \lVert  \tilde{\psi}^{\prime}(C_{i, \cdot} \rho_{\kappa}(\zeta) + d_i)   - \tilde{\psi}^{\prime}(C_{i, \cdot} \rho_{\bar{\kappa}}(\zeta) + d_i) \rVert \lVert (\rho_{\bar{\kappa}}(\zeta))_j \rVert  \\	
		& \quad + \lVert  \tilde{\psi}^{\prime}(C_{i, \cdot} \rho_{\bar{\kappa}}(\zeta) + d_i)  - \tilde{\psi}^{\prime}(\bar{C}_{i, \cdot} \rho_{\bar{\kappa}}(\zeta) + \bar{d}_i)  \rVert \lVert (\rho_{\bar{\kappa}}(\zeta))_j \rVert \big)^2 \\
		& \leq \big( c_1  \lVert (\rho_{\kappa}(\zeta))_j  - (\rho_{\bar{\kappa}}(\zeta))_j\rVert  \\
		& \quad  + c_2 \lVert C_{i, \cdot} \rVert \lVert \rho_{\kappa}(\zeta)   - \rho_{\bar{\kappa}}(\zeta)\rVert \lVert (\rho_{\bar{\kappa}}(\zeta))_j \rVert  \\	
		& \quad  + c_2 (\lVert C_{i, \cdot} -  \bar{C}_{i, \cdot} \rVert \lVert \rho_{\bar{\kappa}}(\zeta) \rVert + \lVert d_i - \bar{d}_i \rVert ) \lVert (\rho_{\bar{\kappa}}(\zeta))_j \rVert \big)^2 \\
		& \leq 3 c_1^2  \lVert (\rho_{\kappa}(\zeta)  - \rho_{\bar{\kappa}}(\zeta))_j \rVert^2  \\
		& \quad   + 3 c_2^2 \lVert C_{i, \cdot} \rVert^2 \lVert \rho_{\kappa}(\zeta)   - \rho_{\bar{\kappa}}(\zeta)\rVert^2 \lVert (\rho_{\bar{\kappa}}(\zeta))_j \rVert^2  \\	
		& \quad  + 3 c_2^2 (\lVert C_{i, \cdot} -  \bar{C}_{i, \cdot} \rVert \lVert \rho_{\bar{\kappa}}(\zeta) \rVert + \lVert d_i - \bar{d}_i \rVert )^2  \lVert (\rho_{\bar{\kappa}}(\zeta))_j \rVert^2,
	\end{split}
	\end{equation*}
\else
	\begin{equation*}
	\begin{split}
	\lVert  \tfrac{\partial}{\partial C_{i,j}}  \chi(\mu) - \tfrac{\partial}{\partial C_{i,j}} \chi(\bar{\mu}) \rVert^2
		& = \lVert \tfrac{\partial}{\partial C_{i,j}} \Psi_{\lambda}(\rho_{\kappa}(\zeta))  - \tfrac{\partial}{\partial C_{i,j}} \Psi_{\bar{\lambda}}(\rho_{\bar{\kappa}}(\zeta)) \rVert^2 \\
		& \leq \big( \lVert  \tilde{\psi}^{\prime}(C_{i, \cdot} \rho_{\kappa}(\zeta) + d_i)\rVert  \lVert (\rho_{\kappa}(\zeta))_j  - (\rho_{\bar{\kappa}}(\zeta))_j\rVert  \\
		& \quad + \lVert  \tilde{\psi}^{\prime}(C_{i, \cdot} \rho_{\kappa}(\zeta) + d_i)   - \tilde{\psi}^{\prime}(C_{i, \cdot} \rho_{\bar{\kappa}}(\zeta) + d_i) \rVert \lVert (\rho_{\bar{\kappa}}(\zeta))_j \rVert  \\	
		& \quad + \lVert  \tilde{\psi}^{\prime}(C_{i, \cdot} \rho_{\bar{\kappa}}(\zeta) + d_i)  - \tilde{\psi}^{\prime}(\bar{C}_{i, \cdot} \rho_{\bar{\kappa}}(\zeta) + \bar{d}_i)  \rVert \lVert (\rho_{\bar{\kappa}}(\zeta))_j \rVert \big)^2 \\
		& \leq \big( c_1  \lVert (\rho_{\kappa}(\zeta))_j  - (\rho_{\bar{\kappa}}(\zeta))_j\rVert  
		  + c_2 \lVert C_{i, \cdot} \rVert \lVert \rho_{\kappa}(\zeta)   - \rho_{\bar{\kappa}}(\zeta)\rVert \lVert (\rho_{\bar{\kappa}}(\zeta))_j \rVert  \\	
		& \quad  + c_2 (\lVert C_{i, \cdot} -  \bar{C}_{i, \cdot} \rVert \lVert \rho_{\bar{\kappa}}(\zeta) \rVert + \lVert d_i - \bar{d}_i \rVert ) \lVert (\rho_{\bar{\kappa}}(\zeta))_j \rVert \big)^2 \\
		& \leq 3 c_1^2  \lVert (\rho_{\kappa}(\zeta)  - \rho_{\bar{\kappa}}(\zeta))_j \rVert^2  
		 + 3 c_2^2 \lVert C_{i, \cdot} \rVert^2 \lVert \rho_{\kappa}(\zeta)   - \rho_{\bar{\kappa}}(\zeta)\rVert^2 \lVert (\rho_{\bar{\kappa}}(\zeta))_j \rVert^2  \\	
		& \quad  + 3 c_2^2 (\lVert C_{i, \cdot} -  \bar{C}_{i, \cdot} \rVert \lVert \rho_{\bar{\kappa}}(\zeta) \rVert + \lVert d_i - \bar{d}_i \rVert )^2  \lVert (\rho_{\bar{\kappa}}(\zeta))_j \rVert^2,
	\end{split}
	\end{equation*}
\fi
where in the last equation we used the Cauchy--Schwarz inequality. 
Summing over $j$, 
yields
\if\twocol1
	\begin{equation*}
	\begin{split}
	\sum_{j=1}^{n_2} & \lVert  \tfrac{\partial}{\partial C_{i,j}}  \chi(\mu) - \tfrac{\partial}{\partial C_{i,j}} \chi(\bar{\mu}) \rVert^2  \\
		& \leq  3 c_1^2  \lVert \rho_{\kappa}(\zeta) - \rho_{\bar{\kappa}}(\zeta) \rVert^2 \\
		&  \quad + 3 c_2^2 \lVert C_{i, \cdot} \rVert^2 \lVert \rho_{\kappa}(\zeta)   - \rho_{\bar{\kappa}}(\zeta)\rVert^2  \lVert \rho_{\bar{\kappa}}(\zeta)\rVert^2  \\	
		&  \quad + 3 c_2^2 \big( \lVert C_{i, \cdot} -  \bar{C}_{i, \cdot} \rVert \lVert \rho_{\bar{\kappa}}(\zeta) \rVert + \lVert d_i - \bar{d}_i \rVert \big)^2  \lVert \rho_{\bar{\kappa}}(\zeta) \rVert ^2  .		 
	\end{split}
	\end{equation*}
\else
	\begin{equation*}
	\begin{split}
	\sum_{j=1}^{n_2} \lVert  \tfrac{\partial}{\partial C_{i,j}}  \chi(\mu) - \tfrac{\partial}{\partial C_{i,j}} \chi(\bar{\mu}) \rVert^2 
		& \leq  3 c_1^2  \lVert \rho_{\kappa}(\zeta) - \rho_{\bar{\kappa}}(\zeta) \rVert^2 
		+ 3 c_2^2 \lVert C_{i, \cdot} \rVert^2 \lVert \rho_{\kappa}(\zeta)   - \rho_{\bar{\kappa}}(\zeta)\rVert^2  \lVert \rho_{\bar{\kappa}}(\zeta)\rVert^2  \\	
		&  \quad + 3 c_2^2 \big( \lVert C_{i, \cdot} -  \bar{C}_{i, \cdot} \rVert \lVert \rho_{\bar{\kappa}}(\zeta) \rVert + \lVert d_i - \bar{d}_i \rVert \big)^2  \lVert \rho_{\bar{\kappa}}(\zeta) \rVert ^2  .		 
	\end{split}
	\end{equation*}
\fi
Summing this expression over $i$, again using the norm \eqref{eq:Euclidean-like norm-Intro}, and using Cauchy--Schwarz for the last term, we get
\if\twocol1
	\begin{equation*}
	\begin{split}
	\lVert &  \tfrac{\partial}{\partial C}  \chi(\mu) - \tfrac{\partial}{\partial C} \chi(\bar{\mu}) \rVert^2  \\
	 	& = \sum_{i=1}^{n_3}\sum_{j=1}^{n_2}  \lVert  \tfrac{\partial}{\partial C_{i,j}}  \chi(\mu) - \tfrac{\partial}{\partial C_{i,j}} \chi(\bar{\mu}) \rVert^2  \\
		& \leq  3 c_1^2 L_1^2 n_3 \lVert \kappa - \bar{\kappa}\rVert^2  \\
		&  \quad +3 c_2^2 \lVert C \rVert^2 L_1^2  \lVert \kappa - \bar{\kappa}\rVert^2  B_1^2  \\	
		&  \quad+ 3 c_2^2 (B_1^2 + 1) \sum_{i=1}^{n_3}\big(  \lVert C_{i, \cdot} -  \bar{C}_{i, \cdot} \rVert^2 + \lVert d_i - \bar{d}_i \rVert^2 \big) B_1^2  \\
		& \leq  3  L_1^2 (c_1^2 n_3 + c_2^2 D^2 B_1^2) \lVert \kappa - \bar{\kappa}\rVert^2  \\
		& \quad  +3 c_2^2 (B_1^2 + 1) B_1^2 \lVert \lambda - \bar{\lambda} \rVert^2 
		.		 
	\end{split}
	\end{equation*}
\else
	\begin{equation*}
	\begin{split}
	\lVert \tfrac{\partial}{\partial C}  \chi(\mu) - \tfrac{\partial}{\partial C} \chi(\bar{\mu}) \rVert^2
	 	& = \sum_{i=1}^{n_3}\sum_{j=1}^{n_2}  \lVert  \tfrac{\partial}{\partial C_{i,j}}  \chi(\mu) - \tfrac{\partial}{\partial C_{i,j}} \chi(\bar{\mu}) \rVert^2  \\
		& \leq  3 c_1^2 L_1^2 n_3 \lVert \kappa - \bar{\kappa}\rVert^2  
		 +3 c_2^2 \lVert C \rVert^2 L_1^2  \lVert \kappa - \bar{\kappa}\rVert^2  B_1^2  \\	
		&  \quad+ 3 c_2^2 (B_1^2 + 1) \sum_{i=1}^{n_3}\big(  \lVert C_{i, \cdot} -  \bar{C}_{i, \cdot} \rVert^2 + \lVert d_i - \bar{d}_i \rVert^2 \big) B_1^2  \\
		& \leq  3  L_1^2 (c_1^2 n_3 + c_2^2 D^2 B_1^2) \lVert \kappa - \bar{\kappa}\rVert^2  
		  +3 c_2^2 (B_1^2 + 1) B_1^2 \lVert \lambda - \bar{\lambda} \rVert^2 
		.		 
	\end{split}
	\end{equation*}
\fi
With a very similar (but slightly easier) computation we get
\if\twocol1
	\begin{equation*}
	\begin{split}
	\lVert &  \tfrac{\partial}{\partial d}  \chi(\mu) - \tfrac{\partial}{\partial d} \chi(\bar{\mu}) \rVert^2  \\
		& \leq  2 c_2^2 \big( D^2 L_1^2 \lVert \kappa - \bar{\kappa} \rVert^2 +  (B_1^2 +1) \lVert \lambda - \bar{\lambda} \rVert^2 \big)
		.		 
	\end{split}
	\end{equation*}
\else
	\begin{equation*}
	\lVert  \tfrac{\partial}{\partial d}  \chi(\mu) - \tfrac{\partial}{\partial d} \chi(\bar{\mu}) \rVert^2  \leq  2 c_2^2 \big( D^2 L_1^2 \lVert \kappa - \bar{\kappa} \rVert^2 +  (B_1^2 +1) \lVert \lambda - \bar{\lambda} \rVert^2 \big) .		 
	\end{equation*}
\fi
Combining these two results we have
\if\twocol1
	\begin{equation} \label{eq:chi-by-lambda}
	\begin{split}
	\lVert &  \tfrac{\partial}{\partial \lambda}  \chi(\mu) - \tfrac{\partial}{\partial \lambda} \chi(\bar{\mu}) \rVert^2  
		\\
		& \leq \big( 3  L_1^2 (c_1^2 n_3 + c_2^2 D^2 B_1^2) +  2 c_2^2 D^2 L_1^2 \big) \lVert \kappa - \bar{\kappa}\rVert^2  \\
		& \quad + c_2^2 (B_1^2 + 1) (3 B_1^2 + 2 ) \lVert \lambda - \bar{\lambda} \rVert^2 \\
		& 
		\leq m_1 \lVert \mu - \bar{\mu} \rVert^2
		.		 
	\end{split}
	\end{equation}
\else
	\begin{equation} \label{eq:chi-by-lambda}
	\begin{split}
	\lVert   \tfrac{\partial}{\partial \lambda}  \chi(\mu) - \tfrac{\partial}{\partial \lambda} \chi(\bar{\mu}) \rVert^2  
		& \leq \big( 3  L_1^2 (c_1^2 n_3 + c_2^2 D^2 B_1^2) +  2 c_2^2 D^2 L_1^2 \big) \lVert \kappa - \bar{\kappa}\rVert^2  \\
		& \quad + c_2^2 (B_1^2 + 1) (3 B_1^2 + 2 ) \lVert \lambda - \bar{\lambda} \rVert^2 \\
		& 
		\leq m_1 \lVert \mu - \bar{\mu} \rVert^2
		.		 
	\end{split}
	\end{equation}
\fi
Now we compute the Lipschitz constant of the last part of the gradient of $\chi$. We proceed similarly to before, using the triangle inequality, yielding
\if\twocol1
	\begin{equation*}
	\begin{split}
	\lVert &  \tfrac{\partial}{\partial \kappa}  \chi(\mu) - \tfrac{\partial}{\partial \kappa} \chi(\bar{\mu}) \rVert^2  \\
		& = \lVert \Psi_{\lambda}^{\prime}(\rho_{\kappa}(\zeta)) \, \nabla \rho_{\kappa}(\zeta) - \Psi_{\bar{\lambda}}^{\prime}(\rho_{\bar{\kappa}}(\zeta)) \, \nabla \rho_{\bar{\kappa}}(\zeta) \rVert^2 \\
		& \leq \big( \lVert \Psi_{\lambda}^{\prime}(\rho_{\kappa}(\zeta)) \rVert  \lVert \nabla \rho_{\kappa}(\zeta) - \nabla \rho_{\bar{\kappa}}(\zeta) \rVert  \\
		& \quad + \lVert \Psi_{\lambda}^{\prime}(\rho_{\kappa}(\zeta)) - \Psi_{\bar{\lambda}}^{\prime}(\rho_{\bar{\kappa}}(\zeta)) \rVert \lVert \nabla \rho_{\bar{\kappa}}(\zeta) \rVert \big)^2 \\
		& \leq \big( n_3 c_1 D L_2 \lVert \kappa - \bar{\kappa} \rVert  \\
		& \quad + B_2 \lVert \Psi_{\lambda}^{\prime}(\rho_{\kappa}(\zeta)) - \Psi_{\bar{\lambda}}^{\prime}(\rho_{\bar{\kappa}}(\zeta)) \rVert \big)^2.
	\end{split}
	\end{equation*}
\else
	\begin{equation*}
	\begin{split}
	\lVert  \tfrac{\partial}{\partial \kappa}  \chi(\mu) - \tfrac{\partial}{\partial \kappa} \chi(\bar{\mu}) \rVert^2
		& = \lVert \Psi_{\lambda}^{\prime}(\rho_{\kappa}(\zeta)) \, \nabla \rho_{\kappa}(\zeta) - \Psi_{\bar{\lambda}}^{\prime}(\rho_{\bar{\kappa}}(\zeta)) \, \nabla \rho_{\bar{\kappa}}(\zeta) \rVert^2 \\
		& \leq \big( \lVert \Psi_{\lambda}^{\prime}(\rho_{\kappa}(\zeta)) \rVert  \lVert \nabla \rho_{\kappa}(\zeta) - \nabla \rho_{\bar{\kappa}}(\zeta) \rVert  
		+ \lVert \Psi_{\lambda}^{\prime}(\rho_{\kappa}(\zeta)) - \Psi_{\bar{\lambda}}^{\prime}(\rho_{\bar{\kappa}}(\zeta)) \rVert \lVert \nabla \rho_{\bar{\kappa}}(\zeta) \rVert \big)^2 \\
		& \leq \big( n_3 c_1 D L_2 \lVert \kappa - \bar{\kappa} \rVert  
		 + B_2 \lVert \Psi_{\lambda}^{\prime}(\rho_{\kappa}(\zeta)) - \Psi_{\bar{\lambda}}^{\prime}(\rho_{\bar{\kappa}}(\zeta)) \rVert \big)^2.
	\end{split}
	\end{equation*}
\fi
We compute the second term as
\if\twocol1
	\begin{equation*}
	\begin{split}
	\lVert & \Psi_{\lambda}^{\prime}(\rho_{\kappa}(\zeta)) - \Psi_{\bar{\lambda}}^{\prime}(\rho_{\bar{\kappa}}(\zeta)) \rVert  \\
		& = \lVert \text{diag}(\psi^{\prime}(C \rho_{\kappa}(\zeta) + d)) C - \text{diag}(\psi^{\prime}(\bar{C} \rho_{\bar{\kappa}}(\zeta) + \bar{d})) \bar{C}\rVert \\
		& \leq \lVert \text{diag}(\psi^{\prime}(C \rho_{\kappa}(\zeta) + d)) \rVert  \lVert C - \bar{C} \rVert  \\
		& \quad +\lVert \text{diag}\big(\psi^{\prime}(C \rho_{\kappa}(\zeta) + d) - \psi^{\prime}(C \rho_{\bar{\kappa}}(\zeta) + d) \big)  \rVert  \lVert \bar{C} \rVert  \\
		& \quad +\lVert \text{diag}\big(\psi^{\prime}(C \rho_{\bar{\kappa}}(\zeta) + d) - \psi^{\prime}(\bar{C} \rho_{\bar{\kappa}}(\zeta) + \bar{d}) \big) \rVert \lVert \bar{C} \rVert \\
		& \leq n_3 c_1  \lVert C - \bar{C} \rVert + c_2 \lVert C \rVert  \lVert \rho_{\kappa}(\zeta) - \rho_{\bar{\kappa}}(\zeta) \rVert  D  \\
		& \quad +D \sqrt{\sum_{i=1}^{n_3} c_2^2 (\lVert C_{i,\cdot} - \bar{C}_{i,\cdot} \rVert \lVert  \rho_{\bar{\kappa}}(\zeta) \rVert + \lVert d_i - \bar{d}_i \rVert )^2 } \\
		& \leq n_3 c_1   \lVert C - \bar{C} \rVert + c_2 D^2 L_1 \lVert \kappa -\bar{\kappa} \rVert   \\
		& \quad +D c_2 \sqrt{B_1^2 + 1} \lVert \lambda - \bar{\lambda} \rVert \\
		& \leq  \big( n_3 c_1  + D c_2 (B_1^2 + 1)^{1/2} \big) \lVert \lambda - \bar{\lambda} \rVert + c_2 D^2 L_1 \lVert \kappa -\bar{\kappa} \rVert,
	\end{split} 
	\end{equation*}
\else
	\begin{equation*}
	\begin{split}
	\lVert \Psi_{\lambda}^{\prime}(\rho_{\kappa}(\zeta)) - \Psi_{\bar{\lambda}}^{\prime}(\rho_{\bar{\kappa}}(\zeta)) \rVert
		& = \lVert \text{diag}(\psi^{\prime}(C \rho_{\kappa}(\zeta) + d)) C - \text{diag}(\psi^{\prime}(\bar{C} \rho_{\bar{\kappa}}(\zeta) + \bar{d})) \bar{C}\rVert \\
		& \leq \lVert \text{diag}(\psi^{\prime}(C \rho_{\kappa}(\zeta) + d)) \rVert  \lVert C - \bar{C} \rVert  \\
		& \quad +\lVert \text{diag}\big(\psi^{\prime}(C \rho_{\kappa}(\zeta) + d) - \psi^{\prime}(C \rho_{\bar{\kappa}}(\zeta) + d) \big)  \rVert  \lVert \bar{C} \rVert  \\
		& \quad +\lVert \text{diag}\big(\psi^{\prime}(C \rho_{\bar{\kappa}}(\zeta) + d) - \psi^{\prime}(\bar{C} \rho_{\bar{\kappa}}(\zeta) + \bar{d}) \big) \rVert \lVert \bar{C} \rVert \\
		& \leq n_3 c_1  \lVert C - \bar{C} \rVert + c_2 \lVert C \rVert  \lVert \rho_{\kappa}(\zeta) - \rho_{\bar{\kappa}}(\zeta) \rVert  D  \\
		& \quad +D \sqrt{\sum_{i=1}^{n_3} c_2^2 (\lVert C_{i,\cdot} - \bar{C}_{i,\cdot} \rVert \lVert  \rho_{\bar{\kappa}}(\zeta) \rVert + \lVert d_i - \bar{d}_i \rVert )^2 } \\
		& \leq n_3 c_1   \lVert C - \bar{C} \rVert + c_2 D^2 L_1 \lVert \kappa -\bar{\kappa} \rVert   
		 +D c_2 \sqrt{B_1^2 + 1} \lVert \lambda - \bar{\lambda} \rVert \\
		& \leq  \big( n_3 c_1  + D c_2 (B_1^2 + 1)^{1/2} \big) \lVert \lambda - \bar{\lambda} \rVert + c_2 D^2 L_1 \lVert \kappa -\bar{\kappa} \rVert,
	\end{split} 
	\end{equation*}
\fi
where we used Cauchy--Schwarz in the second last step and $\lVert C - \bar{C} \rVert \leq \lVert \lambda - \bar{\lambda} \rVert$ in the last step.
Inserting this in the previous inequality yields
\if\twocol1
	\begin{equation}\label{eq:chi-by-kappa}
	\begin{split}
	\lVert &  \tfrac{\partial}{\partial \kappa}  \chi(\mu) - \tfrac{\partial}{\partial \kappa} \chi(\bar{\mu}) \rVert^2  \\
		& \leq \Big( (n_3 c_1 D L_2  + B_2 c_2 D^2 L_1) \lVert \kappa - \bar{\kappa} \rVert  \\
		& \quad + B_2 \big( n_3 c_1  + D c_2 (B_1^2 + 1)^{1/2} \big) \lVert \lambda - \bar{\lambda} \rVert \Big)^2 \\
		& \leq \Big( (n_3 c_1 D L_2  + B_2 c_2 D^2 L_1)^2  \\
		& \quad + B_2^2 \big( n_3 c_1  + D c_2 (B_1^2 + 1)^{1/2} \big)^2 \Big) \lVert \mu - \bar{\mu} \rVert^2 \\
		& = m_2 \lVert \mu - \bar{\mu} \rVert^2.
	\end{split}
	\end{equation}
\else
	\begin{equation}\label{eq:chi-by-kappa}
	\begin{split}
	\lVert \tfrac{\partial}{\partial \kappa}  \chi(\mu) - \tfrac{\partial}{\partial \kappa} \chi(\bar{\mu}) \rVert^2
		& \leq \Big( (n_3 c_1 D L_2  + B_2 c_2 D^2 L_1) \lVert \kappa - \bar{\kappa} \rVert  \\
		& \quad + B_2 \big( n_3 c_1  + D c_2 (B_1^2 + 1)^{1/2} \big) \lVert \lambda - \bar{\lambda} \rVert \Big)^2 \\
		& \leq \Big( (n_3 c_1 D L_2  + B_2 c_2 D^2 L_1)^2  \\
		& \quad + B_2^2 \big( n_3 c_1  + D c_2 (B_1^2 + 1)^{1/2} \big)^2 \Big) \lVert \mu - \bar{\mu} \rVert^2 \\
		& = m_2 \lVert \mu - \bar{\mu} \rVert^2.
	\end{split}
	\end{equation}
\fi
Combining \eqref{eq:chi-by-lambda} and \eqref{eq:chi-by-kappa} we arrive at
\if\twocol1
	\begin{equation*}
	\begin{split}
	\lVert &  \nabla  \chi(\mu) - \nabla \chi(\bar{\mu}) \rVert^2  \\
		& = \lVert \tfrac{\partial}{\partial \kappa}  \chi(\mu) - \tfrac{\partial}{\partial \kappa} \chi(\bar{\mu}) \rVert^2 +  \lVert \tfrac{\partial}{\partial \lambda}  \chi(\mu) - \tfrac{\partial}{\partial \lambda} \chi(\bar{\mu}) \rVert^2  \\
		& \leq (m_1 + m_2) \lVert \mu - \bar{\mu} \rVert^2,
	\end{split}
	\end{equation*}
\else
	\begin{equation*}
	\begin{split}
	\lVert  \nabla  \chi(\mu) - \nabla \chi(\bar{\mu}) \rVert^2
		& = \lVert \tfrac{\partial}{\partial \kappa}  \chi(\mu) - \tfrac{\partial}{\partial \kappa} \chi(\bar{\mu}) \rVert^2 +  \lVert \tfrac{\partial}{\partial \lambda}  \chi(\mu) - \tfrac{\partial}{\partial \lambda} \chi(\bar{\mu}) \rVert^2  \\
		& \leq (m_1 + m_2) \lVert \mu - \bar{\mu} \rVert^2,
	\end{split}
	\end{equation*}
\fi

which proves $ii)$. \\
The second bound in $iii)$ is immediate using the fact that $\chi$ maps to $\R^{n_3}$ and that each component of the resulting vector is bounded by $B_3$. 
For the first bound we remark that the Lipschitz constant is always an upper bound for the gradient, which completes the proof.
\end{proof}

\begin{rem}\label{rem:lem-induction-step}
Under the same setting as in Lemma \ref{lem:induction-step}, but using a function $\psi : \R^{n_3} \to \R$ with constants $c_1, c_2 > 0$ such that for all $x \in \R^{n_3}$ we have $\lVert \tfrac{\partial}{\partial x} \psi(x) \rVert \leq c_1$ and $\lVert \tfrac{\partial^2}{\partial x^2} \psi(x) \rVert \leq c_2$, we get exactly the same constants with $n_3 = 1$. 
Indeed, going through the proof again and replacing $\psi$ wherever necessary, we first get the partial derivatives with $\psi'(x) := \tfrac{\partial}{\partial x} \psi(x)$,
\begin{align}\label{eq:partials-chi-rem}
\tfrac{\partial}{\partial C_{i,j}} \Psi_{\lambda}(x) &= \psi'(Cx + d) x_j e_i ,\\
\tfrac{\partial}{\partial d_i} \Psi_{\lambda}(x) &= \psi'(Cx + d)  e_i, \\
\tfrac{\partial}{\partial \kappa} \chi(\mu) &= \Psi_{\lambda}^{\prime}(\rho_{\kappa}(\zeta)) \, \nabla \rho_{\kappa}(\zeta),\\
\Psi_{\lambda}^{\prime}(x) &=
	\psi^{\prime}(C x + d)) C.
\end{align}
Using them in the subsequent steps, we see that we get exactly the same constants with $n_3 = 1$.
\end{rem}

With Lemma \ref{lem:induction-step} we can now prove Theorems \ref{thm:lipschitz of NN} and \ref{thm:lipschitz of loss} iteratively.

\begin{proof}[Proof of Theorem \ref{rem:lipschitz of NN}]
First, we apply Lemma \ref{lem:induction-step} with $u:=1$, $n_1:=\ell_0$, $n_2:=\ell_{u-1}$, $n_3:=\ell_u$, $(C,d) := \theta_u$,  $\tilde{\psi}:= \tilde{\sigma}_u$, $l_1:=0$, $\rho_{\kappa} := id $ and $\zeta := \zeta_x$. Hence, we have $\tilde{D} :=B_{\Omega}$, $c_1 := \sigma'_{\max}$ and $c_2 := \sigma''_{\max}$, $L_1 := 0$, $L_2 := 0$, $B_1 := S$, $B_2 := 0$, $B_3 := \sigma_{\max}$.  Therefore, $N_1$ and $\nabla N_1$ are  Lipschitz continuous and bounded with the Lipschitz constants $L_{N_1}$, $L_{\nabla N_1}$ and the bounding constants $B_{N_1}$ and $B_{\nabla N_1}$ as given in Theorem \ref{rem:lipschitz of NN}. 
Next, we apply Lemma \ref{lem:induction-step} iteratively, where for $2 \leq u \leq m$ we use the same variables as above except for $l_1 := d_{u-1}$, $\rho_{\kappa} := N_{u-1}$ with $\kappa := \Theta_{u-1}$, yielding $L_1 := L_{N_{u-1}}$, $L_2 := L_{\nabla N_{u-1}}$, $B_1 := B_{N_{u-1}}$, $B_2 := B_{\nabla N_{u-1}}$. 
It follows that $N_u$ and $\nabla N_{u}$ are Lipschitz and bounded with constants
$L_{N_u}$, $L_{\nabla N_u}$, $B_{N_u}$ and $B_{\nabla N_u}$ as in Theorem \ref{rem:lipschitz of NN}.
To get the Lipschitz constants for $N$ and $\nabla N$, we apply  Lemma \ref{lem:induction-step} another time with the same variables for $u:=m+1$, except for $\tilde{\psi} := id$, yielding $c_1 := 1$, $c_2 := 0$ and $B_3 := \infty$. We conclude that $N$ and $\nabla N$ are Lipschitz continuous and that $\nabla N$ is also bounded with the constants given in Theorem \ref{rem:lipschitz of NN}. 
\end{proof}

\begin{proof}[Proof of Corollary \ref{cor:Lip-consts-iteration-solved}]
The first inequalities can easily be proven by induction.
Furthermore, the following inequality can be shown by induction as well.
\if\twocol1
	\begin{equation*}
	\begin{split}
	\gamma_u &:= 2 B_{\nabla N_{u-1}}^2 (\sigma''_{\max})^2 \ell^8 B_{\Omega}^4  L_{N_{u-1}}^2 \\
		& \quad + B_{\nabla N_{u-1}}^2  \ell \big( \sigma'_{\max} + \ell B_{\Omega} \sigma''_{\max} \sqrt{B_{N_{u-1}}^2 + 1} \big)^2,\\
	L_{\nabla N_u}^2 &\leq \left(2 \ell^6 (\sigma_{\max}')^2  B_{\Omega}^2 \right)^{(u-1)} (\sigma_{\max}'')^2 (S^2+1)(3 S^2 + 2) \\
		& \quad + \sum_{k=1}^{u-1} \left(2 \ell^6 (\sigma_{\max}')^2 B_{\Omega}^2 \right)^{(k-1)} (\alpha_{u-k+1} + \gamma_{u-k+1}).
	\end{split}
	\end{equation*}
\else
	\begin{equation*}
	\begin{split}
	\gamma_u &:= 2 B_{\nabla N_{u-1}}^2 (\sigma''_{\max})^2 \ell^8 B_{\Omega}^4  L_{N_{u-1}}^2 
	 + B_{\nabla N_{u-1}}^2 \ell  \big(  \sigma'_{\max} + \ell B_{\Omega} \sigma''_{\max} \sqrt{B_{N_{u-1}}^2 + 1} \big)^2,\\
	L_{\nabla N_u}^2 &\leq \left(2 \ell^6 (\sigma_{\max}')^2 B_{\Omega}^2 \right)^{(u-1)} (\sigma_{\max}'')^2 (S^2+1)(3 S^2 + 2) \\
		& \quad + \sum_{k=1}^{u-1} \left(2 \ell^6 (\sigma_{\max}')^2  B_{\Omega}^2 \right)^{(k-1)} (\alpha_{u-k+1} + \gamma_{u-k+1}).
	\end{split}
	\end{equation*}
\fi
For the equations of $L_{N_u}^2$, the geometric sum equality for $q \neq 1$, $\sum_{k=0}^n q^n = \frac{1 - q^{n+1}}{1 -q}$, can be used to rewrite the sum. Using this together with quite rough approximations, the asymptotic approximation of $L_{\nabla N_{u}}^2$ can be shown.
\end{proof}

\begin{proof}[Proof of Theorem \ref{rem:lipschitz of loss}]
We first prove that for a random variable $Z = (Z_x,Z_y) \sim \P$, the function
\begin{equation*}
\phi : \R^{d_{m+1}} \to \R, \quad \Theta \mapsto \varphi(\Theta, Z),
\end{equation*}
and its gradient $\nabla \phi := \nabla_{\Theta} \phi$ are Lipschitz continuous with integrable constants. To see this, we proceed as in the proof of Theorem \ref{rem:lipschitz of NN} for $1 \leq u \leq m$. Then, to get the Lipschitz constants of $\phi$ and $\nabla \phi$, we apply Lemma \ref{lem:induction-step} as in the final step of the proof of Theorem \ref{rem:lipschitz of NN}, but using $\psi  := g( \cdot, Z_y)$, $c_1 := g'_{\max}$ and $c_2 := g''_{\max}$. With Remark \ref{rem:lem-induction-step} we get the Lipschitz and bounding constants $L_{\phi}$, $L_{\nabla \phi}$ and $B_{\nabla \phi}$ as defined in Theorem \ref{rem:lipschitz of loss}.
From Corollary \ref{cor:Lip-consts-iteration-solved}, we deduce that there exist constants $a_S, b_S \in \R$ such that 
\[
0 \leq L_{\phi}, L_{\nabla \phi}, B_{\nabla \phi} \leq a_S S^2 + b_S.
\] 
Since $\E[S^2] < \infty$, it follows that  $L_{\phi}, L_{\nabla \phi}, B_{\nabla \phi} \in L^1(\P)$.
In the remaining part of the proof we show that we get the constants for $\Phi$ and $\nabla \Phi$ as in Theorem \ref{rem:lipschitz of loss}.
Let $\Theta, \bar{\Theta} \in \Omega$. Then we have by the Lipschitz continuity of $\phi$ that
\begin{equation*}
\begin{split}
\lVert \Phi(\Theta) - \Phi(\bar{\Theta}) \rVert & = \lVert \E[ \varphi(\Theta, Z) ] - \E[ \varphi(\bar{\Theta}, Z) ] \rVert \\
	& \leq \E\left[ \lVert \varphi(\Theta, Z) - \varphi(\bar{\Theta}, Z) \rVert \right] \\
	& 
	=   \E[L_{\phi}] \, \lVert \Theta - \bar{\Theta} \rVert  = L_{\Phi} \lVert \Theta - \bar{\Theta} \rVert
\end{split}
\end{equation*}
Next we remark that 
\begin{equation}\label{eq:gradient-Phi-equality}
\nabla \Phi(\Theta) = \nabla_{\Theta} \E[ \varphi(\Theta, X) ] = \E[ \nabla_{\Theta} \varphi(\Theta, X)],    
\end{equation}
where we used the dominated convergence theorem in the second equality. Indeed, dominated convergence can be used, since  $\nabla_{\Theta} \varphi(\Theta, X)$ exists and 
since all directional derivatives (and sequences converging to them) can be bounded by the following integrable random variable
\if\twocol1
	\[
	\begin{split}
	\lVert \lim_{\epsilon \to 0}\tfrac{1}{\epsilon} & \left( \varphi(\Theta + \epsilon \bar{\Theta}, Z) - \varphi(\Theta, Z) \right) \rVert  \\
		&\leq \lim_{\epsilon \to 0} \tfrac{1}{\epsilon} L_{\phi} \lVert \epsilon \bar{\Theta} \rVert  
		\leq L_{\phi} \sqrt{d_{m+1}} B_{\Omega}.
	\end{split}
	\]
\else
	\[
	\begin{split}
	\lVert \lim_{\epsilon \to 0}\tfrac{1}{\epsilon} & \left( \varphi(\Theta + \epsilon \bar{\Theta}, Z) - \varphi(\Theta, Z) \right) \rVert 
	\leq \lim_{\epsilon \to 0} \tfrac{1}{\epsilon} L_{\phi} \lVert \epsilon \bar{\Theta} \rVert  
		\leq L_{\phi} \sqrt{d_{m+1}} B_{\Omega}.
	\end{split}
	\]
\fi
This also implies that a (vector-valued) sequence converging to the gradient $\nabla_{\Theta} \varphi(\Theta, X)$ can be bounded by an integrable random variable, yielding that the assumptions for dominated convergence are satisfied. 
Hence, we have that
\if\twocol1
	\begin{equation*}
	\begin{split}
	\lVert \nabla \Phi(\Theta) \rVert & = \lVert \E[ \nabla_{\Theta} \varphi(\Theta, Z)] \rVert \leq \E\left[ \lVert \nabla_{\Theta} \varphi(\Theta, Z)  \rVert \right] \\
		& \leq \E[B_{\nabla \phi} ]= B_{\nabla \Phi}
	\end{split}
	\end{equation*}
\else
	\begin{equation*}
	\lVert \nabla \Phi(\Theta) \rVert  = \lVert \E[ \nabla_{\Theta} \varphi(\Theta, Z)] \rVert \leq \E\left[ \lVert \nabla_{\Theta} \varphi(\Theta, Z)  \rVert \right] 
	\leq \E[B_{\nabla \phi} ]= B_{\nabla \Phi}
	\end{equation*}
\fi
and 
\if\twocol1
	\begin{equation*}
	\begin{split}
	\lVert \nabla \Phi(\Theta) &- \nabla \Phi(\bar{\Theta}) \rVert  = \lVert \E[ \nabla_{\Theta} \varphi(\Theta, Z) -  \nabla_{\Theta} \varphi(\bar{\Theta}, Z) ] \rVert \\
		&\leq \E\left[ \lVert \nabla_{\Theta} \varphi(\Theta, Z) -  \nabla_{\Theta} \varphi(\bar{\Theta}, Z) \rVert \right] \\
		& \leq \E[L_{\nabla \phi}] \, \lVert \Theta - \bar{\Theta}  \rVert
		= L_{\nabla \Phi} \lVert \Theta - \bar{\Theta}  \rVert,
	\end{split}
	\end{equation*}
\else
	\begin{equation*}
	\begin{split}
	\lVert \nabla \Phi(\Theta) - \nabla \Phi(\bar{\Theta}) \rVert  
	&= \lVert \E[ \nabla_{\Theta} \varphi(\Theta, Z) -  \nabla_{\Theta} \varphi(\bar{\Theta}, Z) ] \rVert \\
		&\leq \E\left[ \lVert \nabla_{\Theta} \varphi(\Theta, Z) -  \nabla_{\Theta} \varphi(\bar{\Theta}, Z) \rVert \right] \\
		& \leq \E[L_{\nabla \phi}] \, \lVert \Theta - \bar{\Theta}  \rVert
		= L_{\nabla \Phi} \lVert \Theta - \bar{\Theta}  \rVert,
	\end{split}
	\end{equation*}
\fi
which completes the proof.
\end{proof}
\begin{rem}
If $\operatorname{proj}_x(\mathcal{Z})$ is bounded by $B_S>0$, then in Theorem \ref{thm:lipschitz of loss}, $S$ can be chosen to be this bound and we get exactly the same constants, but in this case $L_{\phi}, L_{\nabla \phi}, B_{\nabla \phi}$ are  also constants rather than random variables.
\end{rem}

We can now use Theorem \ref{rem:lipschitz of loss} to prove the two examples.
\begin{proof}[Proof of Example \ref{cor:convergence of GD}]
In this setting of a finite training set with equal probabilities we have for $\Theta \in \Omega$,
\begin{equation*}
\nabla \Phi (\Theta) = \tfrac{1}{N} \sum_{i=1}^N \nabla \varphi(\Theta, \zeta_i).
\end{equation*}
In particular, we can compute the true gradient of $\Phi$. 
By the assumption $\sup_{j \geq 0} \lVert \Theta^{(j)}\rVert_{\infty} < B_{\Omega}$, we can use $\Omega = \lbrace \Theta \in \R^{d_{m+1}} \; | \; \lVert \Theta \rVert_{\infty} < B_{\Omega} \rbrace$. 
Furthermore, since the training set is finite (and hence bounded), we can set $S:= \max_{1 \leq i \leq N} \lVert \zeta_i \rVert < \infty$, and get by Theorem \ref{rem:lipschitz of loss} that $\Phi$ and $\nabla \Phi$ are Lipschitz continuous on $\Omega$ with constants $L_{\Phi}$ and $L_{\nabla \Phi}$.
The result then follows as outlined in Section 1.2.3 of \cite{nesterov2013introductory}. 
\end{proof}
\begin{proof}[Proof of Example \ref{cor:convergence of SGD}]
By the assumption $\sup_{j \geq 0} \lVert \Theta^{(j)}\rVert_{\infty} < B_{\Omega}$, we can use $\Omega = \lbrace \Theta \in \R^{d_{m+1}} \; | \; \lVert \Theta \rVert_{\infty} < B_{\Omega} \rbrace$. 
%
Furthermore, since $S = \lVert \operatorname{proj}_x(Z) \rVert \in L^2$, Theorem \ref{rem:lipschitz of loss} 
yields, that $\Phi$ and $\nabla \Phi$ are Lipschitz continuous on $\Omega$ with constants $L_{\Phi}$ and $L_{\nabla \Phi}$. 
We establish the assumptions of Theorem 4 in \cite{Li_Orabona_2019}, which in turn establishes our result. 
We set $f := \Phi$ and remark first that their results still hold when restricting $\Phi$ and $\nabla \Phi$ to be Lipschitz only on the subset $\Omega$. Indeed, by the assumption $\sup_{j \geq 0} \lVert \Theta^{(j)}\rVert_{\infty} < B_{\Omega}$ we know that $\Theta$ stays within $\Omega$ for the entire training process.
In the remainder of the proof we show that all needed assumptions \textbf{H1}, \textbf{H3} and \textbf{H4'} (as defined in \cite{Li_Orabona_2019}) are satisfied.
\textbf{H1}, the Lipschitz continuity of $\nabla \Phi$, holds as outlined above. 
Let $Z_1, \dotsc, Z_M \sim \P$ be independent and identically distributed random variables with the distribution of the training set.
By the stochastic gradient method outlined in \eqref{alg:gradient-method}, in each step the approximation of the gradient $\nabla \Phi(\Theta^{(j)})$ is given by the random variable 
\begin{equation*}
G_j := G(\Theta^{(j)}; Z_1, \dotsc, Z_M) :=  \tfrac{1}{M} \sum_{i=1}^M \nabla_{\Theta} \varphi(\Theta^{(j)}, Z_i).
\end{equation*}
By \eqref{eq:gradient-Phi-equality} we have $\E[ G_j ] = \nabla \Phi$, yielding \textbf{H3}.\\
In the proof of Theorem 4 of \cite{Li_Orabona_2019}, assumption \textbf{H4'} is only used for the proof of their Lemma 8. In particular, it is only used to show
\begin{equation}\label{eq:proof-corollary-SGD-H4-condition}
\E\left[\max_{1 \leq i \leq T} \lVert \nabla \Phi(\Theta_i) - G_i \rVert^2 \right] \leq \sigma^2 (1 + \log(T)),
\end{equation}
for a constant $\sigma \in \R$.
Instead of showing \textbf{H4'}, we directly show that \eqref{eq:proof-corollary-SGD-H4-condition} is satisfied.
We have 
\if\twocol1
	\begin{equation*}
	\begin{split}
	\E&\left[  \max_{1 \leq i \leq T} \lVert \nabla \Phi(\Theta_i) - G_i \rVert^2 \right] \\
		& \leq \E\left[ \max_{1 \leq i \leq T} \left( 2 \lVert \nabla \Phi(\Theta_i) \rVert^2 + 2 \lVert G_i \rVert^2 \right) \right] \\
		& \leq 2 B_{\nabla \Phi}^2 + 2 \E\left[ \max_{1 \leq j \leq T} \tfrac{1}{M} \sum_{i=1}^M \lVert \nabla_{\Theta} \varphi(\Theta^{(j)}, Z_i) \rVert^2  \right] \\
		& \leq  2 B_{\nabla \Phi}^2 + 2 \E[ B_{\nabla \phi}^2] =: \sigma^2,
	\end{split}
	\end{equation*}
\else
	\begin{equation*}
	\begin{split}
	\E\left[  \max_{1 \leq i \leq T} \lVert \nabla \Phi(\Theta_i) - G_i \rVert^2 \right]
		& \leq \E\left[ \max_{1 \leq i \leq T} \left( 2 \lVert \nabla \Phi(\Theta_i) \rVert^2 + 2 \lVert G_i \rVert^2 \right) \right] \\
		& \leq 2 B_{\nabla \Phi}^2 + 2 \E\left[ \max_{1 \leq j \leq T} \tfrac{1}{M} \sum_{i=1}^M \lVert \nabla_{\Theta} \varphi(\Theta^{(j)}, Z_i) \rVert^2  \right] \\
		& \leq  2 B_{\nabla \Phi}^2 + 2 \E[ B_{\nabla \phi}^2] =: \sigma^2,
	\end{split}
	\end{equation*}
\fi
where in the second inequality we used Cauchy--Schwarz and in the last equality we used that $\E[ B_{\nabla \phi}^2] < \infty$, since $S \in L^2$.
In particular this implies that \eqref{eq:proof-corollary-SGD-H4-condition} is satisfied. For completeness we also remark that \textbf{H2} holds as well, since $\Phi$ is Lipschitz. Applying Theorem 4 of \cite{Li_Orabona_2019} concludes the proof.
\end{proof}

\textbf{Constant $C$}

Actually, the adaptive step-sizes $h_j$ of the stochastic gradient method in Algorithm \ref{alg:gradient-method}  can be chosen as
\begin{equation*}
h_j := \frac{\alpha}{\left( \beta + \sum_{i=1}^{j-1} \lVert G_i \rVert^2 \right)^{\frac{1}{2}}}\,.
\end{equation*}
for constants $\alpha, \beta > 0$ that satisfy $2 \alpha L_{\nabla \Phi} <  \sqrt{\beta}$.
We made the choice of taking $\alpha = 1$ and  $\beta = 4 L_{\nabla \Phi}^2+\varepsilon$ for some $\varepsilon >0$, but this is only one possibility. 
For general $\alpha$ and $\beta$ the constant $C$ is
\[
C = \max( 2\gamma, \sqrt{2\gamma}(\beta+2n\sigma^2)^{1/4}),
\] 
with $\gamma = O\left( \frac{1+\alpha^2\ln n}{\alpha( 1- \tfrac{2\alpha}{\sqrt{\beta}})}\right)$ and $\sigma^2 =2 L_{\Phi}^2 + 2 \E[ B_{\nabla \phi}^2]$. More precisely, the constant $\gamma$ can  explicitly be written as
\begin{equation}
\gamma = \frac{1}{\alpha\left(1-\tfrac{2\alpha L_{\nabla \Phi}}{\sqrt{\beta}}\right)} \left( \Phi(\Theta^{(0)}) - \Phi^*  + \frac{2 \alpha^2 L_{\nabla \Phi}}{\sqrt{\beta}}\sigma^2\right) + K,
\end{equation}
where 
$ K = D\ln(2A, 32B^4D^2+2B^2C+8B^3D\sqrt{C}) = O\left(\frac{\ln (n+1)}{1-\frac{2\alpha L_{\nabla \Phi}}{\beta}}\right)$ with 
\begin{eqnarray*}
A &=& \sqrt{\beta+2(n+1)\sigma^2}, \qquad B=\sqrt{2}, \qquad D=\tfrac{\alpha n}{1-\tfrac{2\alpha n}{\sqrt{\beta}}}\\
C &=& \frac{\beta\left(\Phi(\Theta^{(0)}) - \Phi^*\right)+2\alpha(1+\ln n)\sigma^2}{\alpha\beta\left(1-\frac{2\alpha L_{\nabla \Phi}}{\sqrt{\beta}}\right)}
\end{eqnarray*}
For more details, see Theorem 4 and its proof of \cite{Li_Orabona_2019}.

\section{Bounding $\Omega$ with 2-norm instead of $\infty$-norm}\label{sec: Bounding Omega with 2-norm instead of infty-norm}
	
If we bound $\lambda$ in Lemma \ref{lem:induction-step} by $\lVert \lambda \rVert < \tilde{D}$ instead of $\lVert \lambda \rVert_{\infty} < \tilde{D}$, then by setting $D := \tilde{D}$, the claims of the Lemma hold equivalently.
Therefore, Theorem~~\ref{thm:lipschitz of NN} and Theorem~~\ref{thm:lipschitz of loss} hold equivalently, when assuming that for all $\Theta \in \Omega$ we have $\lVert \Theta \rVert < B_{\Omega}$ and setting  $D_u := B_{\Omega}$.

\textbf{Improvement of the upper bounds on  the Lipschitz constants.}
Bounding the 2-norm of the parameters $\Theta$ implies that in each layer the weights can only contribute to a certain amount of the total norm. Therefore, an improvement of the constants is possible in this setting.

In particular, revisiting the proof of Theorem \ref{thm:lipschitz of NN} we see that in each step applying Lemma \ref{lem:induction-step}, we set $D = B_\Omega$. However, in the $u$-th layer, $D$ needs only to be a bound for $\lVert \theta_u \rVert$, while the norm of the entire parameter vector has to satisfy $\lVert \Theta_{m+1} \rVert < B_{\Omega}$. 
Therefore, we can replace $D_u = \sqrt{\ell_{u_1} \ell_u} B_\Omega$ by  $D_u \geq 0$ in the constants for the $u$-th layer in Theorem \ref{thm:lipschitz of NN}. Doing this for all layers, including the last one, $L_N$ and $L_{\nabla N}$ become functions of $(D_1,\dotsc,D_{m+1})$, where the constraint $\sum_{u=1}^{m+1}D_u^2 < B_\Omega^2$ has to be satisfied.
Hence, computing tighter upper bounds of the Lipschitz constants amounts to solving the optimization problems
\if\twocol1
	\begin{align*}
	&\max\left\{ L_N(D_1, \dotsc, D_{m+1})  \Big\vert \sum_{u=1}^{m+1}D_u^2 < B_\Omega^2 \right\}, \\
	&\max\left\{ L_{\nabla N}(D_1, \dotsc, D_{m+1})  \Big\vert \sum_{u=1}^{m+1}D_u^2 < B_\Omega^2 \right\}.
	\end{align*}
\else
	\begin{equation*}
	\max\left\{ L_N(D_1, \dotsc, D_{m+1})  \Big\vert \sum_{u=1}^{m+1}D_u^2 < B_\Omega^2 \right\} \; \text{and} \;
	\max\left\{ L_{\nabla N}(D_1, \dotsc, D_{m+1})  \Big\vert \sum_{u=1}^{m+1}D_u^2 < B_\Omega^2 \right\}.
	\end{equation*}
\fi
Due to the iterative definition of $L_N$ and $L_{\nabla N}$, both objective functions are complex polynomials in a high dimensional constraint space where the maximum is achieved at some boundary point, i.e. where $\sum_{u=1}^{m+1} D_u^2 = B_\Omega^2$.
In particular, numerical methods have to be used to solve these optimization problems.

\section{Auxiliary results in the controlled ODE setting}\label{sec:Auxiliary results in the controlled ODE setting}

\begin{example}\label{exa:CODE-NN equivalence}
We define $u$ as a step function and $V^{\theta}$ as a stepwise (with respect to its second parameter) vector field
\if\twocol1
	\begin{equation*}
	\begin{split}
	u(t) &:= \sum_{i=1}^{m} i \, \mathbbm{1}_{[i, i+1)} (t) + (m+1) \, \mathbbm{1}_{[m+1, \infty)}(t) \\
	V^{\theta} (t,x) &:= \sum_{i=1}^m \mathbbm{1}_{(i-1, i]}(t) \left( \sigma_i(f_{\theta_i}(x)) - x \right) \\
		& \quad \quad + \mathbbm{1}_{(m, \infty)}(t) \left( f_{\theta_{m+1}}(x) - x) \right.
	\end{split}
	\end{equation*}
\else
	\begin{equation*}
	\begin{split}
	u(t) &:= \sum_{i=1}^{m} i \, \mathbbm{1}_{[i, i+1)} (t) + (m+1) \, \mathbbm{1}_{[m+1, \infty)}(t) \\
	V^{\theta} (t,x) &:= \sum_{i=1}^m \mathbbm{1}_{(i-1, i]}(t) \left( \sigma_i(f_{\theta_i}(x)) - x \right) + \mathbbm{1}_{(m, \infty)}(t) \left( f_{\theta_{m+1}}(x) - x) \right.
	\end{split}
	\end{equation*}
\fi
Here, $\theta = (\theta_1, \dotsc, \theta_{m+1})$ is the concatenation of all the  weights needed to define the affine neural network layers, and $\sigma_i$ and $f_{\theta_i}$ are defined as in Section \ref{sec:Problem set-up}.
However, by abuse of notation, we assume that each $f_{\theta_i}: \R^{\ell} \to \R^{\ell}$, using ``$0$-embeddings'' wherever needed and similar for $\sigma_i$.
Evaluating \eqref{eq:solved controlled ODE}, which amounts to computing the (stochastic) integral with respect to a step function, we get
\begin{equation*}
\begin{split}
X_t^{\theta,x} & = x + \sum_{i=1}^{m+1} 1_{\lbrace i \leq t \rbrace} V^{\theta}(i, X^{\theta,x}_{i^{-}}) \left( u(i) - u(i-)  \right) \\
	& = x + \sum_{i=1}^{m+1} 1_{\lbrace i \leq t \rbrace} \left( \sigma_i(f_{\theta_i}(X_{i^{-}}^{\theta,x})) - X_{i^{-}}^{\theta,x} \right),
\end{split}
\end{equation*}
where we use $\sigma_{m+1} := id$. Solving the sum iteratively, we get for $1 \leq i \leq m+1 =: T$,
\if\twocol1
	\begin{equation*}
	\begin{split}
	X^{\theta, x}_0 &= x, \\
	X^{\theta, x}_i &= \sigma_i \circ f_{\theta_i} \circ \dotsb \circ \sigma_1 \circ f_{\theta_1} (x), 
	\end{split}
	\end{equation*}
\else
	\begin{equation*}
	\begin{split}
	X^{\theta, x}_0 &= x, \quad
	X^{\theta, x}_i = \sigma_i \circ f_{\theta_i} \circ \dotsb \circ \sigma_1 \circ f_{\theta_1} (x), 
	\end{split}
	\end{equation*}
\fi
in particular, $X^{\theta,x}_i$ is the output of the $i^{\text{th}}$ layer of the neural network $\mathcal{N}_{\Theta_{m+1}}$ defined in \eqref{eq:NN definition}.
\end{example}

\begin{rem}\label{rem:Schwarz Thm}
If $\partial_x V_i^{\theta}(t, x)$ is continuously differentiable with respect to $\theta$, then Schwarz's theorem, as for example outlined in Chapter 2.3 of \cite{konigsberger2013analysis}, implies that $\partial_{\theta} \partial_x V_i^{\theta}(t, x) = \partial_x \partial_{\theta} V_i^{\theta}(t, x)$. In particular, the bounding constants $B_{\partial_{x \theta} V}$ and $B_{\partial_{ \theta x} V}$ are equal.
\end{rem}

\begin{rem}\label{rem:compareing theorems}
Comparing the theorems of Section \ref{sec:Deep neural networks as controlled ODEs} to the theorems of Section \ref{sec:Ordinary deep neural network setting}, we see that here we did not make assumptions on the boundedness of $\Omega$. 
As we discussed before, the controlled ODE setting \eqref{eq:solved controlled ODE} is a generalization of the setting in Section \ref{sec:Ordinary deep neural network setting}, hence, Theorem \ref{thm:Lipschitz continuity of X} and \ref{thm:Lipschitz continuity of Phi - controlled ODE} can be applied to a classical DNN. 
Does this mean that the assumption of $\Omega$ being bounded is in fact unnecessary? The answer is no, because for the assumptions \eqref{eq:assump thm Lipschitz continuity of X} on the vector fields $V_i$ to be satisfied in the case of DNN, it is necessary to assume that $\Omega$ is bounded. In that sense, this assumption is now just hidden inside another assumption.\\
Furthermore, it is easy to see that the constants estimated in Theorem \ref{thm:lipschitz of NN} and \ref{thm:lipschitz of loss} are smaller than the respective constants that we get from Theorem \ref{thm:Lipschitz continuity of X} and \ref{thm:Lipschitz continuity of Phi - controlled ODE}.
\end{rem}

\section{Proofs in the controlled ODE setting}\label{sec:Proofs in the controlled ODE setting}
For the proofs of the Lipschitz results we extensively use the following stochastic version of \emph{Gr\"onwall's Inequality}, which is presented as Lemma 15.1.6 in \cite{cohen2015stochastic}.

\begin{lem}\label{lem:Gronwall}
Let $Y$ be a (1-dimensional) c\`adl\`ag process, $U$ an increasing real process and $\alpha > 0$ a constant. If for all $0 \leq t \leq T$, 
\[
Y_t \leq \alpha + \int_0^t Y_{s-} dU_s,
\]
then $Y_t \leq \alpha \mathcal{E}(U)_t$ for all $0 \leq t \leq T$.
\end{lem}
Here $\mathcal{E}(U)$ is the stochastic exponential as defined in Definition 15.1.1 and Lemma 15.1.2 of \cite{cohen2015stochastic}. Note also that $0 \leq \mathcal{E}(U)_t \leq \exp(U_t)$ holds, if no jump of $U$ is smaller than $-1$, i.e. $\Delta U_s \geq -1$ for all $0 \leq s \leq t$.

\begin{rem}
If $\partial_x V_i^{\theta}(t, x)$ is continuously differentiable with respect to $\theta$, then Schwarz's theorem, as for example outlined in Chapter 2.3 of \cite{konigsberger2013analysis}, implies that $\partial_{\theta} \partial_x V_i^{\theta}(t, x) = \partial_x \partial_{\theta} V_i^{\theta}(t, x)$. In particular, the bounding constants $B_{\partial_{x \theta} V}$ and $B_{\partial_{ \theta x} V}$ are equal.
\end{rem}

\begin{proof}[Proof of Theorem \ref{thm:Lipschitz continuity of X}]
Starting from \eqref{eq:solved controlled ODE} we get
\begin{equation*}
\begin{split}
\lVert X_{t}^{\theta} \rVert & \leq  \lVert x \rVert + \left\lVert \sum_{i=1}^d \int_{0}^t V_{i}^{\theta}\left(s,X_{s-}^{\theta}\right)du_{i}(s) \right\rVert \\
	& \leq \lVert x \rVert +  \int_{0}^t \max_{1 \leq i \leq d} \lVert V_{i}^{\theta}\left(s,X_{s-}^{\theta}\right) \rVert d\upsilon(s) \\
	& \leq \lVert x \rVert +  \int_{0}^t B_V ( 1 + \lVert X_{s-}^{\theta} \rVert) d\upsilon(s) \\
	& = \lVert x \rVert +  B_V B_\upsilon + \int_{0}^t  \lVert X_{s-}^{\theta} \rVert d\tilde{\upsilon}(s),
\end{split}
\end{equation*}
where $\tilde{\upsilon} = B_V \upsilon$.
Hence, Lemma \ref{lem:Gronwall} implies that for $0 \leq t \leq T$,
\begin{equation*}
\begin{split}
\lVert X_{t}^{\theta} \rVert 
	& \leq (\lVert x \rVert + B_V B_\upsilon)\mathcal{E}(B_V \upsilon)_t \\
	&\leq (\lVert x \rVert + B_V B_\upsilon) \exp(B_V B_\upsilon) = B_X,
\end{split}
\end{equation*}
using the fact that all the jumps of $u$ are positive, since $u$ is increasing.
In the following we do the same for \eqref{ODE} and \eqref{eq:ODE2}, showing that the first and second derivatives of $X_{T}^{\theta}$ with respect to $\theta$ are bounded, which implies that $\theta \mapsto X_{T}^{\theta}$ and $\theta \mapsto \partial X_{T}^{\theta}$ are Lipschitz continuous on $\Omega$ with these constants.
Using all the given bounds and using that $\lVert \partial_x V_i \rVert$ is bounded by the Lipschitz constant $L_{V_x}$, we obtain  from \eqref{ODE} the following inequality:
\begin{equation*}
\begin{split}
\lVert \partial X_{t}^{\theta} \rVert \leq  B_{\partial_{\theta} V} (1+B_X^{p_{\theta}}) B_\upsilon 
	+ \int_0^t L_{V_x} \lVert \partial X_{s-}^{\theta} \rVert d\upsilon(s)
\end{split}.
\end{equation*}
Hence, by Lemma \ref{lem:Gronwall}, we have for $0 \leq t \leq T$,
\begin{equation*}
\lVert \partial X_{t}^{\theta} \rVert \leq B_{\partial_{\theta} V} (1+B_X^{p_{\theta}}) B_\upsilon \exp(L_{V_x} B_\upsilon) = L_X,
\end{equation*}
which therefore is a Lipschitz constant of the map $\theta \mapsto X_{T}^{\theta}$. Similarly, we need the corresponding ODE for the second derivative of $X_{t}^{\theta}$  with respect to $\theta$ in order to obtain the Lipschitz constant of the map $\theta \mapsto \partial X_{T}^{\theta}$. Assuming that all needed derivatives of $V_i^{\theta}$ exist, similarly to \eqref{ODE}, we obtain the ODE for the second derivative
\if\twocol1
	\begin{equation}
	\label{eq:ODE2}
	\begin{split}
	d\partial_{\theta \theta} X_{t}^{\theta} &= \displaystyle \sum_{i=1}^d \left[ \partial_{\theta \theta} V_{i}^{\theta}\left(t, X_{t-}^{\theta}\right) \right.\\ 
	      & \qquad \quad  + \left. \partial_{x \theta} V_{i}^{\theta}\left(t,X_{t-}^{\theta}\right) \partial X_{t-}^{\theta} \right. \\
	      &\qquad \quad  + \left. \partial_{\theta x} V_{i}^{\theta}\left(t,X_{t-}^{\theta}\right) \partial X_{t-}^{\theta} \right. \\
	      &\qquad \quad + \left. {\partial_{x x} V_{i}^{\theta}\left(t,X_{t-}^{\theta}\right) \partial X_{t-}^{\theta}} \partial X_{t-}^{\theta} \right. \\
	      &\qquad \quad + \left. \partial_{x} V_{i}^{\theta}\left(t,X_{t-}^{\theta}\right) \partial_{\theta \theta} X_{t-}^{\theta} 
	      \right] du_{i}(t), \\
	      \partial_{\theta \theta} X_{0}^{\theta} &= 0 \in \mathbb{R}^{\ell \times n \times n}.
	\end{split}
	\end{equation}
\else
	\begin{equation}
	\label{eq:ODE2}
	\begin{split}
	d\partial_{\theta \theta} X_{t}^{\theta} &= \displaystyle \sum_{i=1}^d \left[ \partial_{\theta \theta} V_{i}^{\theta}\left(t, X_{t-}^{\theta}\right) 
		 +  \partial_{x \theta} V_{i}^{\theta}\left(t,X_{t-}^{\theta}\right) \partial X_{t-}^{\theta} 
		   +  \partial_{\theta x} V_{i}^{\theta}\left(t,X_{t-}^{\theta}\right) \partial X_{t-}^{\theta} \right. \\
	      &\qquad \quad + \left. {\partial_{x x} V_{i}^{\theta}\left(t,X_{t-}^{\theta}\right) \partial X_{t-}^{\theta}} \partial X_{t-}^{\theta} 
	       +  \partial_{x} V_{i}^{\theta}\left(t,X_{t-}^{\theta}\right) \partial_{\theta \theta} X_{t-}^{\theta} 
	      \right] du_{i}(t), \\
	      \partial_{\theta \theta} X_{0}^{\theta} &= 0 \in \mathbb{R}^{\ell \times n \times n}.
	\end{split}
	\end{equation}
\fi
Here, we have an equation for tensors of third order. We implicitly assume that for each term the correct tensor product is used, such that the term has the required dimension. Writing down the equation component wise clarifies which tensor products are needed. Observe that \eqref{eq:ODE2} is also a linear ODE, and therefore, by Theorem 7 of Chapter V in \cite{Pro1992}, a unique solution exists. 
Finally, for \eqref{eq:ODE2} we get
\begin{align*}
&\lVert \partial_{\theta \theta} X_{t}^{\theta} \rVert \leq C_{\theta \theta} + \int_0^t L_{V_x} \lVert \partial_{\theta \theta} X_{s-}^{\theta} \rVert d\upsilon(s).
\end{align*}
Hence, by Lemma \ref{lem:Gronwall}, we have for $0 \leq t \leq T$,
\begin{equation*}
\lVert \partial_{\theta \theta} X_{t}^{\theta} \rVert \leq C_{\theta \theta} \exp(L_{V_x} B_\upsilon) = L_{\partial X},
\end{equation*}
which is therefore a Lipschitz constant of $\theta \mapsto \partial X_{T}^{\theta}$.
\end{proof}

\begin{proof}[Proof of Theorem \ref{thm:Lipschitz continuity of Phi - controlled ODE}]
We first prove that for a random variable $ Z = (Z_x, Z_y) \sim \P$, the function
\begin{equation*}
\phi: \Omega \to \R, \theta \mapsto \phi(\theta) := \varphi(\theta, Z),
\end{equation*}
and its gradient $\nabla \phi := \nabla_{\Theta} \phi$ are Lipschitz continuous with integrable constants.
In the following, $L_X$ and $L_{\partial X}$ are as defined in the proof of Theorem \ref{thm:Lipschitz continuity of X}, except that $\lVert x \rVert$ is now exchanged with $S$ (in the definition of $B_X$). 
Let $\theta, \bar{\theta} \in \Omega$. Then
\begin{equation*}
\begin{split}
\lVert \phi(\theta)  - \phi(\bar{\theta}) \rVert &= \lVert g(X_T^{\theta, Z_x}, Z_y) - g(X_T^{\bar\theta, Z_x}, Z_y)\rVert \\
	& \leq L_g L_X \lVert \theta - \bar{\theta} \rVert,
\end{split}
\end{equation*}
which shows the first part of the claim. We define $L_{\phi} := L_g L_X$ and also notice that, by Lipschitz continuity of $\phi$, it follows that the gradient $\nabla \phi$ is bounded by $L_{\phi}$.
Furthermore, we have
\if\twocol1
	\begin{equation*}
	\begin{split}
	\lVert &\nabla \phi (\theta) - \nabla \phi (\bar\theta) \rVert \\
		& = \lVert \partial_x g  (X_T^{\theta, Z_x} , Z_y) \partial X_T^{\theta, Z_x} - 
			  \partial_x g (X_T^{\bar\theta, Z_x} , Z_y) \partial X_T^{\bar\theta, Z_x} \rVert \\
		& \leq (L_{\partial_x g} L_X^2 + L_g L_{\partial X}) \lVert \theta - \bar{\theta}  \rVert,
	\end{split}
	\end{equation*}
\else
	\begin{equation*}
	\begin{split}
	\lVert \nabla \phi (\theta) - \nabla \phi (\bar\theta) \rVert
		& = \lVert \partial_x g  (X_T^{\theta, Z_x} , Z_y) \partial X_T^{\theta, Z_x} - 
			  \partial_x g (X_T^{\bar\theta, Z_x} , Z_y) \partial X_T^{\bar\theta, Z_x} \rVert \\
		& \leq (L_{\partial_x g} L_X^2 + L_g L_{\partial X}) \lVert \theta - \bar{\theta}  \rVert,
	\end{split}
	\end{equation*}
\fi
where we used again the technique to introduce intermediate terms and split up the norm using the triangle inequality. Defining $L_{\nabla \phi} := (L_{\partial_x g} L_X^2 + L_g L_{\partial X})$, this shows the second part of the claim. 
Since $\E[S^p] < \infty$, it follows that  $L_{\phi}, L_{\nabla \phi} \in L^1(\P)$.\\
Using this and the same steps as in the second part of the proof of Theorem \ref{thm:lipschitz of loss}, we now get that $\Phi$ and $\nabla \Phi$ are Lipschitz continuous with constants $L_{\Phi} = \E[L_{\phi}]$ and $L_{\nabla\Phi} = \E[L_{\nabla\phi}]$ and that $\nabla \Phi$ is bounded by $B_{\nabla \Phi} := L_{\Phi}$. 
\end{proof}

\fi

\end{document}